%% file: main.tex
\documentclass[10pt,twocolumn,letterpaper]{article}

\usepackage{cvpr}              

\input{preamble}

\definecolor{cvprblue}{rgb}{0.21,0.49,0.74}
\usepackage[pagebackref,breaklinks,colorlinks,citecolor=cvprblue]{hyperref}
\usepackage{lipsum}
\usepackage{array}
\usepackage{caption}
\usepackage{booktabs}
\usepackage[accsupp]{axessibility}
\usepackage{xspace}
\usepackage{multirow}
\usepackage{pifont}
\usepackage{enumitem}
\usepackage{colortbl}

\def\paperID{222} 
\def\confName{3DV\xspace}
\def\confYear{2026\xspace}

\newcommand{\methodname}{DyTact\xspace}
\newcommand{\datasetname}{\textit{\methodname-$21$}\xspace}
\newcommand{\RomanNumeralCaps}[1]
    {\MakeUppercase{\romannumeral #1}}
\title{DyTact: Capturing Dynamic Contacts in Hand-Object Manipulation}

\author{Xiaoyan Cong$^1$ \quad Angela Xing$^1$ \quad    Chandradeep Pokhariya$^2$ \quad  Rao Fu$^1$ \quad Srinath Sridhar$^1$\thanks{Corresponding author}\\
$^1$Brown University  \qquad $^2$IIT Delhi \\
}

\begin{document}
\maketitle
\input{sec/0_abstract}

\input{sec/1_intro}
\input{sec/2_short_related_work}
\input{sec/3_preliminary}

\input{sec/4_method}
\input{sec/5_experiment}

\input{sec/7_conclusion}

\paragraph{\textbf{Acknowledgements.}}
This work was supported by NSF CAREER grant \#2143576, ONR DURIP grant N00014-23-1-2804, a gift from Meta Reality Labs, and an AWS Cloud Credits award.

{
    \small
    \bibliographystyle{ieeenat_fullname}
    \bibliography{main}
}

\end{document}

%% file: preamble.tex
\usepackage[dvipsnames]{xcolor}


%% file: sec/0_abstract.tex
\begin{abstract}
\noindent
Reconstructing dynamic hand-object contacts is essential for realistic manipulation in AI character animation, XR, and robotics, yet it remains challenging due to heavy occlusions, complex surface details, and limitations in existing capture techniques. 
In this paper, we introduce \methodname, a markerless capture method for accurately capturing dynamic contact in hand–object manipulations in a non-intrusive manner. 
Our approach leverages a dynamic, articulated representation based on 2D Gaussian surfels to model complex manipulations. 
By binding these surfels to MANO~\cite{MANO:SIGGRAPHASIA:2017} meshes, \methodname harnesses the inductive bias of template models to stabilize and accelerate optimization. 
A refinement module addresses time-dependent high-frequency deformations, while a contact-guided adaptive sampling strategy selectively increases surfel density in contact regions to handle heavy occlusion. 
Extensive experiments demonstrate that \methodname not only achieves state-of-the-art dynamic contact estimation accuracy but also significantly improves novel view synthesis quality, all while operating with fast optimization and efficient memory usage.
\end{abstract}

%% file: sec/1_intro.tex
\section{Introduction}
Skillful object manipulation is one of the most common, yet impressive, human physical abilities.
Human manipulation of objects is highly dynamic, and often bimanual, involving the coordinated movements of fingers in both hands to perform complex tasks.
An important step in analyzing or replicating manipulations is understanding the \textbf{dynamic contacts} between hands and objects~\cite{elguea2023review}.
Contact not only provides a measure of spatial proximity~\cite{zhang2021manipnet, brahmbhatt2019contactdb, taheri2020grab} but also conveys object affordances~\cite{jian2023affordpose,yang2022oakink,wu2023learning}.
Moreover, variations in contact over time influence both the kinematics and dynamics~\cite{beltran2020learning,zhang2023efficient,liu2025parameterized} of the interaction.

\begin{figure}[t]
  \centering
  \includegraphics[width=1\linewidth]{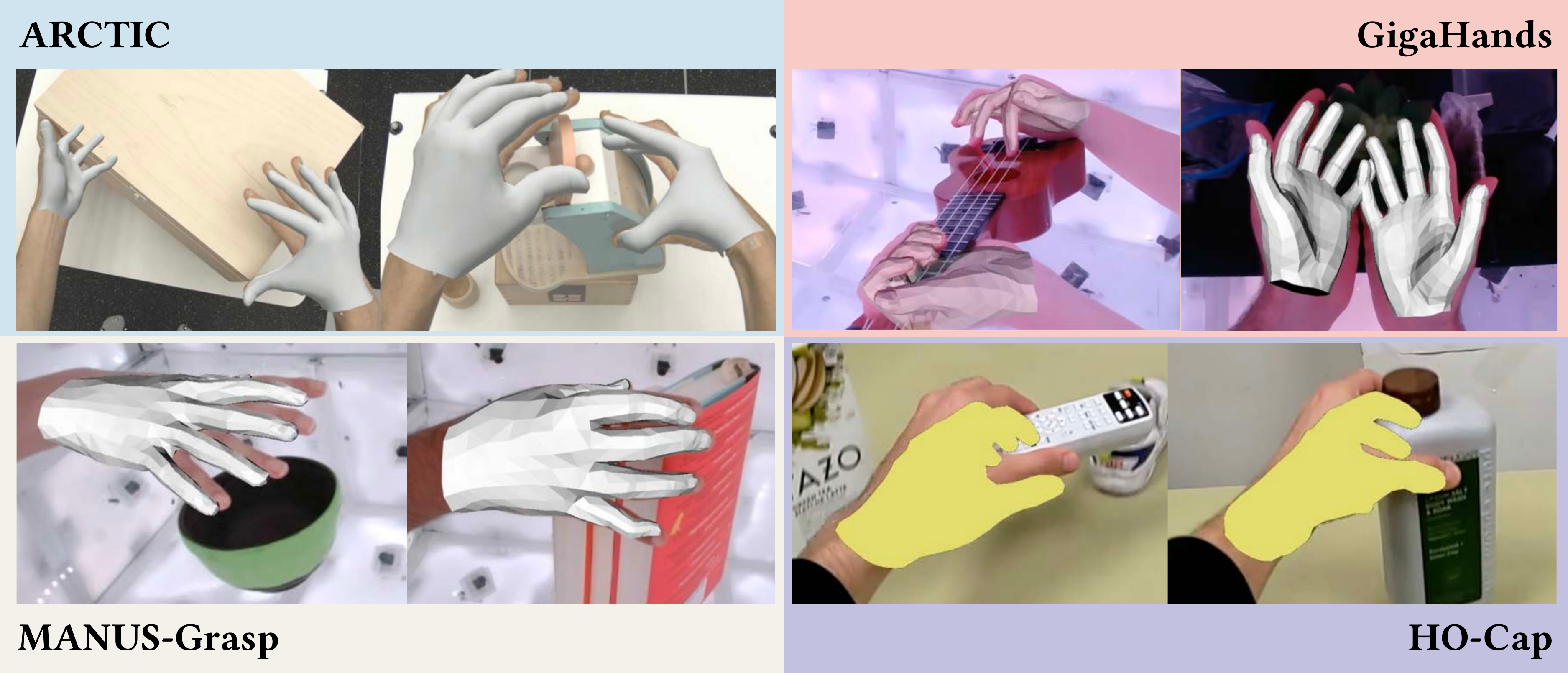}
  \vspace{-0.1in}
  \caption{We demonstrate the \textbf{misalignment} between the parametric hand shape templates and actual hands by overlapping the rendering of actual hands with the annotated ``Ground Truth'' MANO meshes from four datasets, ARCTIC~\cite{fan2023arctic}, GigaHands~\cite{fu2024gigahandsmassiveannotateddataset}, MANUS-Grasp~\cite{pokhariya2024manus}, HO-Cap~\cite{wang2025hocapcapturedataset3d}.
  Such misalignments exacerbate errors in dynamic contact estimation.
  }
  \vspace{-0.1in}
  \label{Fig:Misalignment}
\end{figure}

Despite its significance, accurately capturing and reconstructing dynamic contacts remains a challenge.
Traditional hardware-based approaches find it challenging to capture dynamic contacts. 
For example, instrumented gloves~\cite{heumer2007grasp, lin2014grasp, sundaram2019learning} can be intrusive, as they might constrain natural movements and affect tactile feedback, which compromises the realism and accuracy of capturing. 
Thermal sensors~\cite{brahmbhatt2019contactdb} can only estimate the accumulated contact in an interaction sequence. 
Therefore, they cannot sensitively capture instantaneous contact during dynamic manipulation.
Moreover, most hardware solutions are expensive and difficult to scale up, limiting their potential applications.

\begin{figure*}
  \includegraphics[width=\textwidth]{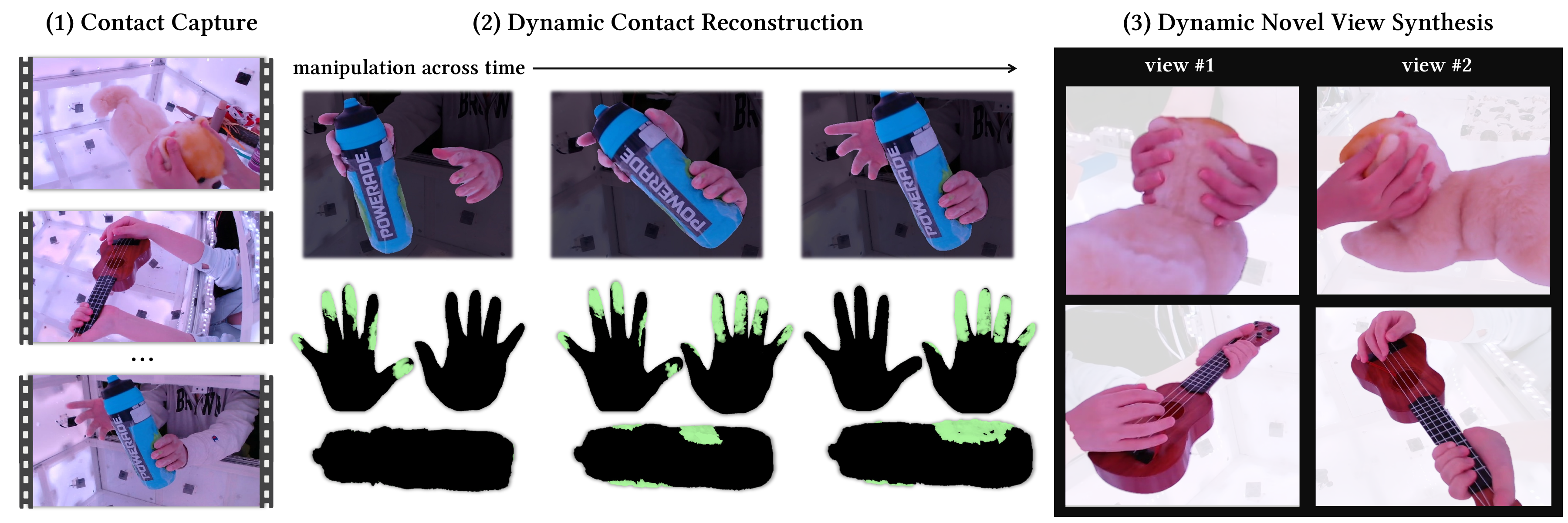}
  \vspace{-0.25in}
    \caption{ \textbf{\methodname} is a markerless \underline{Dy}namic con\underline{Tact} capture method for complex hand-object manipulations.
    \methodname employs markerless contact capture (left) and uses a dynamic articulated representation based on 2D Gaussians to accurately model hand-object contacts without misalignments.
    Experimental results demonstrate that \methodname achieves accurate dynamic contact estimations (middle), and high-fidelity novel view synthesis (right).
    }
  \vspace{-0.15in}

    \label{fig:teaser}
\end{figure*}

Recently, markerless capture~\cite{gall2009motion, brahmbhatt2020contactpose, liu2024taco, chao2021dexycb, hampali2020honnotate, sridhar2016real, pokhariya2024manus} capable of estimating contact from visual inputs have emerged as a popular and affordable approach. 
These non-intrusive methods enable natural motion during capture and facilitate the subsequent reconstruction of geometry and appearance.
Neural-hand~\cite{karunratanakul2023harp, mundra2023livehand, pokhariya2024manus} provides a data-driven markerless contact capture approach that requires vast amounts of diverse data to train a personalized hand avatar, which is inefficient and not user-friendly.
Another prominent class of solutions leverages parametric hand shape templates, exemplified by MANO~\cite{MANO:SIGGRAPHASIA:2017}. 
Although MANO-based approaches~\cite{liu2024easyhoi, wang2025hocapcapturedataset3d, fan2023arctic, yang2022oakink, zhan2024oakink2, fu2024gigahandsmassiveannotateddataset, ye2022hand, banerjee2024hot3d, chao2021dexycb} exhibit strong generalizability across most scenarios. However, they frequently suffer from noticeable misalignments between the parametric template and the actual hand geometry, as shown in Fig.~\ref{Fig:Misalignment}, which undermines the fidelity of fine-grained contact estimation.
We argue that this limitation stems from the fact that MANO provides a generic, low-dimensional representation of hand shape, whereas accurate contact estimation  necessitates a personalized representation.
Additionally, MANO pose estimation often lacks sufficient robustness and accuracy due to the presence of frequent occlusions and rapid motion during manipulation, thereby exacerbating errors in dynamic contact estimation.


To address the aforementioned limitations, we introduce \textbf{\methodname}, which captures \underline{Dy}namic Con\underline{Tact} for complex hand-object manipulations. 
Our method is non-intrusive, accurate, efficient, and effective with occlusion. It employs multi-view markerless capture for natural manipulations, representing hands and objects with 2D Gaussian surfels~\cite{Huang2DGS2024} which accurately model surfaces and appearance, enabling accurate contact analysis while inherently avoiding the misalignment issues associated with purely parametric templates.
Specifically, for each hand, we bind 2D Gaussian surfels to a parametric hand mesh~\cite{MANO:SIGGRAPHASIA:2017} and optimize a refinement module to handle time-dependent complex surface deformation, ensuring fine detail alignment.
Unlike grasping scenarios where the object typically remains static~\cite{pokhariya2024manus}, our approach supports dynamic manipulation involving both in-hand and between-hand object motions. 
To facilitate this, we initialize a model-free 2D Gaussian surfel representation for the object and track its pose over time. 
Our explicit Gaussian-based representation allows for the efficient computation of both \emph{instantaneous contacts} (defined as contacts at specific frames) and \emph{accumulated contacts} (contacts aggregated over sequences), based on surfel pair distances. 
Furthermore, to address frequent occlusions, we introduce a contact-guided adaptive density control strategy, which selectively prunes surfels while maintaining accurate alignment with a minimal number of surfels.


In summary, our contributions are:
\begin{itemize}[itemsep=1.5pt, topsep=3pt]
\item We introduce \methodname, a method for accurate \underline{Dy}namic con\underline{Tact} capture in complex hand-object manipulation.
%
\item \methodname reconstructs \textbf{both the hand and the object} with dynamic 2D Gaussian surfels~\cite{Huang2DGS2024}, enabling high-fidelity surface modeling without misalignments. 
We propose a contact-guided adaptive density control strategy to effectively address self-occlusions and object occlusions, as well as a time-dependent refinement module that precisely captures complex surface deformations for accurate contact estimation and dynamic reconstruction.
%
\item Experimental results demonstrate the superior performance of \methodname in accurate dynamic contact estimation and high-fidelity novel view synthesis, coupled with fast optimization and efficient memory usage. 
The code and benchmark will be made publicly available upon acceptance.
\end{itemize}

%% file: sec/2_short_related_work.tex
\section{Related Work}
\textbf{Capturing and Modeling Contact.}
Traditional methods have relied on instrumented gloves~\cite{heumer2007grasp, lin2014grasp, sundaram2019learning}, specialized sensors~\cite{yuan2017bighand22mbenchmarkhandpose, pham2017hand, garciahernando2018firstpersonhandactionbenchmark}, or thermal imaging~\cite{brahmbhatt2019contactdb} to capture contacts. 
However, these hardware-based approaches face difficulties in capturing dynamic contacts effectively. 
For instance, instrumented gloves~\cite{heumer2007grasp, lin2014grasp, sundaram2019learning} can be intrusive, as they may constrain natural movements and affect tactile feedback. 
Thermal sensors~\cite{brahmbhatt2019contactdb} cannot capture dynamic or instantaneous contact. 
Moreover, a common bottleneck for most hardware solutions is their high cost and difficulty to scale, limiting their widespread application.

Recent markerless motion capture approaches~\cite{gall2009motion, brahmbhatt2020contactpose, liu2024taco, chao2021dexycb, hampali2020honnotate, sridhar2016real, pokhariya2024manus} are capable of estimating contacts from visual inputs, but the accuracy and efficiency remain significant hurdles.
For example, neural hand avatars~\cite{karunratanakul2023harp, mundra2023livehand, pokhariya2024manus} require extensive training time to fit a personalized model from vast amounts of sequences, which is inefficient and not user-friendly for scalable deployment.

Another prominent class of solutions leverages parametric hand shape templates MANO~\cite{MANO:SIGGRAPHASIA:2017}. 
Although MANO-based approaches~\cite{liu2024easyhoi, wang2025hocapcapturedataset3d, fan2023arctic, yang2022oakink, zhan2024oakink2, fu2024gigahandsmassiveannotateddataset, ye2022hand, banerjee2024hot3d, chao2021dexycb} exhibit good generalizability in many scenarios, they often suffer from significant misalignments between the parametric template and the actual hand (as shown in Fig.~\ref{Fig:Misalignment}). 
These misalignments hinder the capture of fine-grained contact details, compromising the accuracy of contact estimation.
Benefiting from the pixel-wise supervision, \methodname faithfully reconstructs both hands and objects without misalignments.

\textbf{Dynamic Scene Representation.}
Neural representations have recently become a prominent paradigm for scene modeling, achieving considerable progress in novel view synthesis. 
Buiding on Neural Radiance Fields (NeRFs)~\cite{mildenhall2020nerf} and 3D Gaussian Splatting (3D-GS)~\cite{kerbl3Dgaussians}, a significant body of work on modeling dynamic scenes has emerged~\cite{xu20244k4d, attal2023hyperreel, cao2023hexplane, wang2023mixed, lin2023high, huang2024sc, kratimenos2025dynmf, lin2024gaussian, luiten2023dynamic, yang2024deformable, wu20244d, xie2024physgaussianphysicsintegrated3dgaussians, yang2023real, duan20244d}.
While these techniques perform well in general scenarios, few offer a dedicated approach for articulated objects such as hands. 
Efforts like GaussianAvatars~\cite{qian2024gaussianavatars} and SurFhead~\cite{lee2024surfhead} rig Gaussians to FLAME~\cite{li2017learning} for head avatar reconstruction.
However, exploiting dynamic reconstruction for complex hand-object manipulations with MANO~\cite{MANO:SIGGRAPHASIA:2017} remains largely unexplored. 
This paper introduces the first MANO-based 2D Gaussian hand representation, achieving accurate dynamic contact capture and high-fidelity novel view synthesis for complex manipulation scenarios.

We defer the detailed related work to the supplementary material.

%% file: sec/3_preliminary.tex
\begin{figure*}[t]
  \centering
  \includegraphics[width=0.95\linewidth]{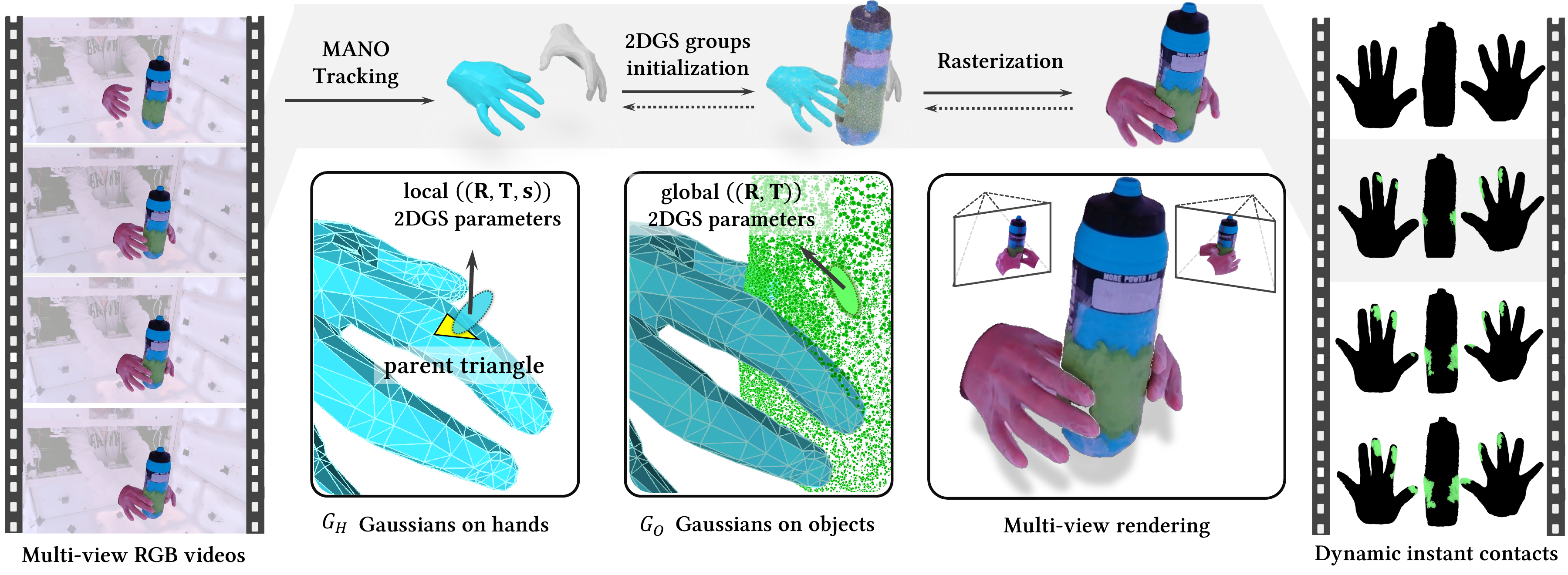}
  \vspace{-0.05in}
  
  \caption{\textbf{\methodname} captures dynamic contacts with a markerless system using a surface-aware dynamic articulated Gaussian representation. 
  Given multi-view RGB videos, it initializes Gaussian surfels on the hand by binding them to the tracked MANO mesh locally, which remain rigged throughout optimization. 
  For objects, Gaussians are initialized by placing a coarse point cloud in the global coordinate space. 
  A refinement module addresses time-dependent high-frequency deformations.
  A contact-guided adaptive sampling strategy selectively refines surfel density in contact regions to handle heavy occlusion, and further optimization of the surfels’ geometry and appearance parameters ensures high-fidelity reconstructions that enable accurate contact estimation.
  }
  \label{Fig:Pipeline}
\vspace{-0.1in}
\end{figure*}

\section{Preliminary}
\label{Sec:Preliminary}

\paragraph{Gaussian Splatting.} 
Given multi-view images and a sparse point cloud of a scene, 3D Gaussian Splatting (3D-GS)~\cite{kerbl3Dgaussians} represents the scenes as a collection of anisotropic 3D Gaussians primitives $G = \{\mathbf{g}_i\}_{i=1}^N$. The geometry of each 3D Gaussian primitive is characterized by a mean position $\mathbf{x}_i \in \mathbb{R}^3$ and a 3D covariance matrix $\mathbf{\Sigma}_i$:
\begin{equation}
    \mathbf{g}_i(\mathbf{x}) = \text{exp}(-\frac{1}{2}(\mathbf{x} - \mathbf{x}_i)^T\mathbf{\Sigma}_i(\mathbf{x} - \mathbf{x}_i)) \nonumber
\label{eq::3DGS_geo}
\end{equation}
where the covariance matrix $\mathbf{\Sigma}_i$ is decomposed into a rotation matrix $\mathbf{R}_i$ and a scaling matrix $\mathbf{S}_i$ as $\mathbf{\Sigma}_i = \mathbf{R}_i\mathbf{S}_i\mathbf{S}_i^T\mathbf{R}_i^T$. $\mathbf{R}_i$ and $\mathbf{S}_i$ are stored as a rotation quaternion $\mathbf{r}_i \in \mathbb{R}^4$ and a scaling factor $\mathbf{s}_i \in \mathbb{R}^3$ respectively for independent optimization. The appearance of each 3D Gaussian is modeled by an opacity value $\sigma_i \in \mathbb{R}$ and a color $\mathbf{c}_i \in \mathbb{R}^k$ defined via $k$ spherical harmonics (SH) coefficients. Overall, $G = \{\mathbf{g}_i = \{\mathbf{x}_i, \mathbf{r}_i, \mathbf{s}_i, \sigma_i, \mathbf{c}_i\}\}_{i=1}^N$. 
The properties of Gaussians are optimized by minimizing the rendering loss between the synthesized and reference images.
To render an image, a tile-based differentiable rasterizer projects Gaussians onto the image plane and applies standard $\alpha$-blending.

2D Gaussian Splatting (2D-GS)~\cite{Huang2DGS2024} extends 3D-GS to model more accurate geometry reconstruction by changing 3D Gaussian primitives to 2D Gaussian surfels. Each 2D Gaussian surfel is geometrically defined by its mean position $\mathbf{x}$, scaling $\mathbf{s}=(s_u, s_v)$, and rotation $\mathbf{r} = (r_u, r_v)$, where $r_u$ and $r_v$ and two principal tangential vectors and $s_u$ and $s_v$ control the corresponding variances. 
Ray-splat Intersection is used in the splatting process to enhance surface modeling quality.
In this paper, we adopt 2DGS as our preferred representation to more accurately model geometry surfaces for contact analysis.


%% file: sec/4_method.tex
\section{\methodname}
\label{Sec:Method}
Given multi-view videos $\mathcal{V} = \{ \mathcal{V}_i:\{\mathbf{I}_i^j \} | 1\leq i \leq N, 1\leq j \leq T \}$ with $N$ views, $T$ frames, and their camera parameters, \methodname accurately reconstructs the geometry, appearance, and contacts between hands and objects in a manipulation sequence.
We achieve this by representing the dynamic scene using time-varying surface geometry and analytically estimating contacts. Specifically, we optimize separate sets of 2D Gaussian surfels for both hands and objects to enable detailed and efficient modeling.
Figure \ref{Fig:Pipeline} presents our pipeline. In this section,
we first describe our data pre-processing and initialization steps in Section~\ref{subsec:initialization}. 
We then introduce a template-based Gaussian representation for hands in Section~\ref{subsec:hand_rep} with time-dependent deformation refinement.
Section~\ref{subsec:scene_compose} describes how we represent the object and how we compose the hand and object to model the entire scene. 
Given our hand and object representations, Section~\ref{par:contact_estimation} describes the dynamic contact estimation process.
Finally, we detail our optimization strategy with contact-guided adaptive density control handling occlusions in Section~\ref{subsec:optimization}. 

\subsection{Initialization}
\label{subsec:initialization}
To accurately capture hand–object manipulations, we extract clear and consistent hand and object masks from the input multi-view images.
We employ Segment Anything V2~\cite{ravi2024sam2} to obtain the foreground segmentation masks $\mathcal{M}=\{ \mathcal{M}_i^j | 1\leq i \leq N, 1\leq j \leq T \}$, which include both hands and objects.
Facilitating the precise geometry and appearance modeling, we initialize a coarse hand surface representation using the MANO model~\cite{romero2022embodied}. A fully automated pipeline~\cite{fu2024gigahandsmassiveannotateddataset} estimates a sequence of MANO meshes $\mathcal{T}=\{\mathcal{T}^j:\{\theta^j, \beta^j, R^j,T^j \} | 1\leq j \leq T\}$ from the input multi-view videos $\mathcal{V}$ to initialize the hand(s), where $\theta$, $\beta$, $R$, and $T$ denote the pose, shape, relative rotation, and translation, respectively.
Similarly, we initialize each object's geometry using coarse point clouds, $\mathcal{O}$, obtained either from offline scans or from the reconstruction of the first frame. 
In summary, our inputs consist of $\{ \mathcal{V}, \mathcal{M}, \mathcal{T}, \mathcal{O}\}$ .

\subsection{Template-based Gaussian Hand}
\label{subsec:hand_rep}
%
To accurately capture hand surface geometry and appearance, we attach 2D Gaussian surfels to the triangular faces of the MANO mesh. 
Each surfel is defined in the local coordinate system of its parent triangle rather than moving freely in 3D space. 
With the MANO parameters $\mathcal{T}^j$ at time step $j$, the dynamics of each surfel are decoupled into a global transformation—driven by the parent triangle’s motion in world coordinates—and a relative transformation within the triangle’s local system. 
This decoupling introduces an inductive bias that significantly accelerates the convergence of Gaussian fitting.

Following the approach in~\cite{qian2024gaussianavatarsphotorealisticheadavatars}, we define each triangle’s local coordinate system by setting its barycenter as the origin $\mathbf{T}$. 
We then construct a rotation matrix $\mathbf{R}$ by concatenating the direction vector of one edge, the triangle’s normal vector, and their cross product. This matrix transforms coordinates from the triangle’s local system to the global coordinate system. 
In the local system, each 2D Gaussian surfel is characterized by a mean position $\mathbf{x}$, rotation $\mathbf{r}$, and anisotropic scaling $\mathbf{s}$. 
These attributes are transformed to the world coordinates as follows:
%
\begin{align}
(\mathbf{r'},\,\mathbf{x'},\,\mathbf{s'})
=
\bigl(\mathbf{R}\mathbf{r},\;s\,\mathbf{R}\mathbf{x} + \mathbf{T},\;s\,\mathbf{s}\bigr),
\end{align}
%
%
%
where $s$ is an isotropic scale representing the triangle's area. 
This scale factor is crucial for speeding up and stabilizing the optimization process, as it adjusts the step size relative to the triangle's area, where larger triangles naturally allow for larger step sizes.

As one single surfel per triangle is insufficient to capture the subtle interactions between hands and objects,
to improve contact estimation capability, we randomly initialize $k$ 2D Gaussian surfels within each triangle's local coordinate system by sampling from a Gaussian distribution centered at $\mathbf{T}$ with variance $v$. 
Empirically, we set $k=5$ and $v=0.5$.  
The left hand and right hand are represented by two separate groups of such 2D Gaussian surfels as $\mathcal{G}_H =\{ \mathcal{G}_H^{\text{left}}, \mathcal{G}_H^{\text{right}}\}$. 
With the binding relationship, our template-based Gaussian representation can be driven by any sets of MANO parameters.

\paragraph{Time-dependent Hand Deformation Refinement.} Although such template-based Gaussian representation can model approximate dynamics, some time-dependent high-frequency deformations might be challenging to capture accurately, such as the complex deformations of the hand skin near the contacting region. 
To address this, we introduce a refinement module $\mathcal{R}_{\theta}$ that compensates these time-dependent high-frequency deformations. 
Taking the attributes $\{\mathbf{x'}, \mathbf{r'}, \mathbf{s'}\}$ of a Gaussian surfel in the world coordinate, and the corresponding timestamp $j$ as inputs, the refinement module $\mathcal{R}_{\theta}$ outputs the offsets of that Gaussian surfel's attributes $ \{ \delta \mathbf{x'}, \delta \mathbf{r'}, \delta \mathbf{s'}\}$ in the world coordinate:
\begin{align}
    (\delta \mathbf{x'}, \delta \mathbf{r'}, \delta \mathbf{s'}) = \mathcal{R}_{\theta}(&\gamma (SG(\mathbf{x'}), L_x), \gamma (SG(\mathbf{r'}), L_r), \nonumber \\  &\gamma (SG(\mathbf{s'}), L_s), \gamma (j, L_j)),
\end{align}
where $\mathcal{R}_{\theta}$ is parameterized as an 8-Layer MLP with hidden dimension 256, $SG(\cdot)$ represents the step-gradiant operation, $\gamma(\cdot, L)$ is the same positional encoding as~\cite{mildenhall2020nerf}. 
Empirically, we set $L_x=8$, $L_r = 4$, $L_s = 4$, $L_j = 4$. 
Then the mean position, rotation and scaling in the world coordinate can be updated as $\{\mathbf{x'} + \delta \mathbf{x'}, \mathbf{r'} + \delta \mathbf{r'}, \mathbf{s'} + \delta \mathbf{s'}\}$ respectively.

\subsection{Object Representation and Scene Composition}
\label{subsec:scene_compose}
%
%
Given a sparse point cloud as initialization, we represent an object by a group of 2D Gaussian surfels $\{\mathbf{g}_i = \{\mathbf{x}_i, \mathbf{r}_i, \mathbf{s}_i, \sigma_i, \mathbf{c}_i\}\}_{i=1}^N$ in the world coordinate. 
We introduce a learnable parameter $\mathbf{P}=\{\mathbf{q}, \mathbf{t}\}$ to track the object's pose along the sequences, where the quaternion $\mathbf{q}$ and vector $\mathbf{t}$ indicate the rotation and translation respectively. 
Thus, all the objects can be represented as $\mathcal{G}_O = \{ \mathcal{G}_O^{o}|o=1,2,\cdots\}$, where $\mathcal{G}_O^{o} = \{ \{\mathbf{g}_i\}_{i=1}^N, \{\mathbf{P}^j\}_{j=1}^T\} \}$.
The entire dynamic scene can be composed as $\mathcal{G} = \{\mathcal{G}_H, \mathcal{G}_O \}$. 
At timestep $j$, hands and objects are driven to the deformed space by the corresponding MANO parameters $\mathcal{T}^j$ and pose parameters $\mathbf{P}^j$. 
During rendering, all the 2D Gaussian surfels are projected onto an image plane and rendered by a differentiable tile-based rasterizer.
During training, we adopt adaptive density control with binding inheritance~\cite{qian2024gaussianavatarsphotorealisticheadavatars} for hands and regular adaptive density control~\cite{Huang2DGS2024} for objects.

\subsection{Dynamic Contact Estimation}
\label{par:contact_estimation}
%
Leveraging our accurate hand and object surface models, we estimate hand–object contact at each frame by comparing their respective Gaussian surfels, following prior analytical methods~\cite{pokhariya2024manus, taheri2020grab, fan2023arctic}.
Specifically, for each 2D Gaussian surfel on the hand, we find the closest 2D Gaussian surfel on the object. 
This pair is considered to be in contact if the distance is less than a pre-defined threshold $\tau$. Given a group of hand Gaussian surfels $\mathcal{G}_H$ and object Gaussian surfels $\mathcal{G}_O$, the contact map between them is defined as:
\begin{equation}
    \text{C} = \begin{cases} 
\mathbf{D}(\mathbf{g}_H, \mathbf{g}_O) & \text{if } \mathbf{D}(\mathbf{g}_H, \mathbf{g}_O) < \tau, \\
0 & \text{otherwise}.
\end{cases}
\end{equation}
where $\mathbf{g}_H \in \mathcal{G}_H$, $\mathbf{g}_O \in \mathcal{G}_O$, and $\mathbf{D}$ represents the Euclidean distance between positions of two 2D Gaussian surfels.

\begin{figure}[t]
  \centering
  \includegraphics[width=0.99\linewidth]{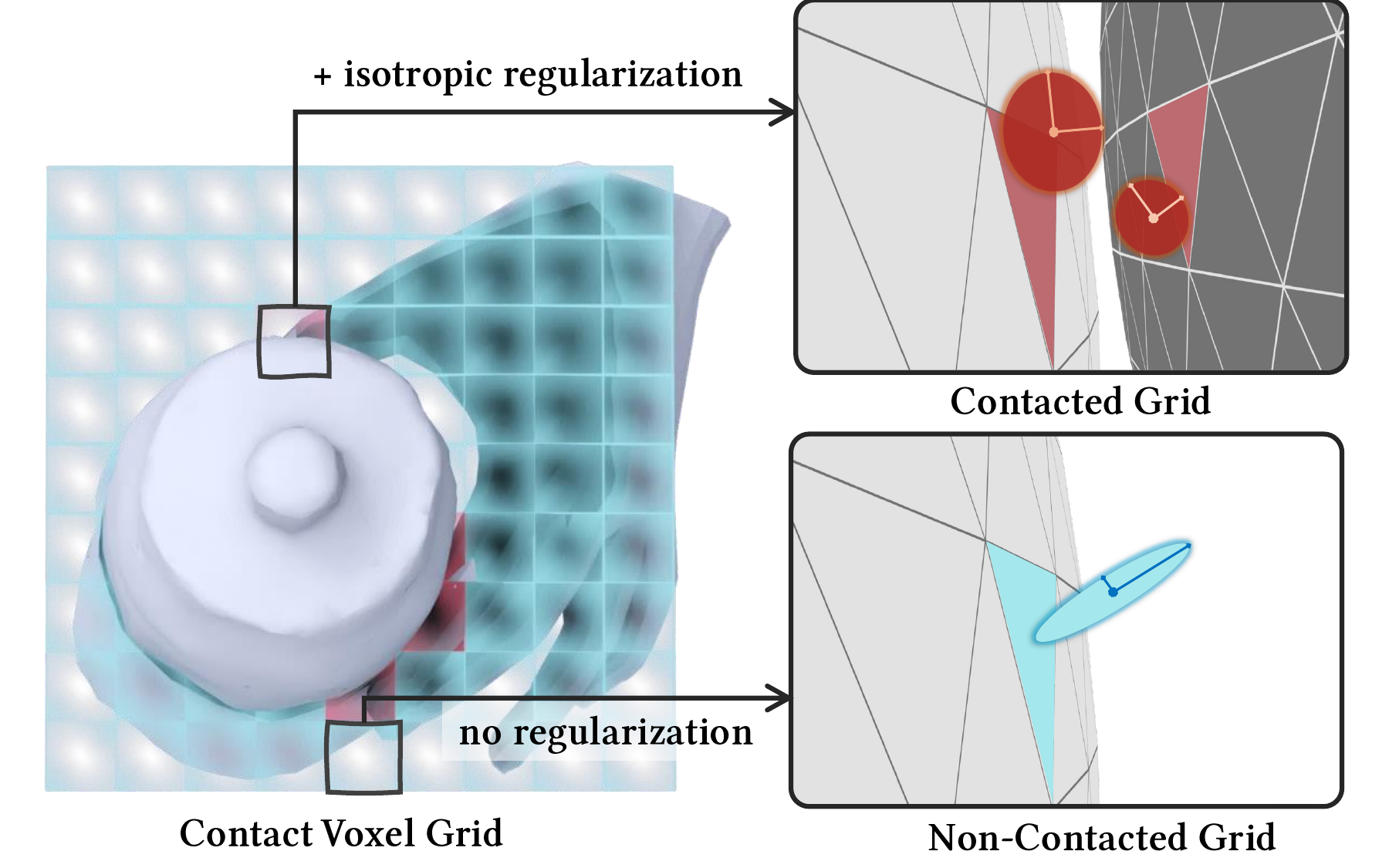}
  \vspace{-0.05in}
  \caption{\textbf{Contact-guided adaptive sampling strategy} ensures Gaussians remain well-regularized in occluded areas.
  The entire space was voxelized and labeled based on contact status. An isotropic regularization is applied to Gaussian surfels within contacting regions, effectively preserving and accumulating informative gradients around contacting regions.}
  \vspace{-0.1in}
  \label{Fig:CGADCS}
\end{figure}

\subsection{Optimization}
\label{subsec:optimization}

\paragraph{Contact-Guided Adaptive Density Control.}
To model regions with varying complexity more efficiently, adaptive density control strategy~\cite{kerbl3Dgaussians} provides a general guidance: intuitively, it allocates more Gaussian primitives to regions with higher complexity and vice versa. 
In hand-object manipulation scenes, self-occlusions and hand-object occlusions where interactions happen are more complicated and harder to fit. 
We observe that Gaussian surfels with weird shapes such as abnormal narrow long disks cause artifacts in the rendered contact maps. 
To capture more accurate contacts, we propose a contact-guided adaptive density control strategy with motivations of allocating more Gaussian surfels around the contacting regions and making them as isotropic as possible. 
As shown in Figure \ref{Fig:CGADCS}, We first voxelize the whole scene with a voxel size $\tau_v$. 
We set $\tau_v = \tau / \sqrt{3}$ and $\tau$ is the pre-defined contact threshold in Section~\ref{par:contact_estimation}. 
A voxel is identified as \textit{contact-voxel} when it contains both hand Gaussian surfels and object Gaussian surfels. 
Then we introduce an isotropic regularization term $\mathcal{L}_i$ to constrain the variances of two orthogonal axes of Gaussian surfels within \textit{contact-voxels}:
\begin{equation}
    \mathcal{L}_i = || \frac{\text{min}_{\mathbf{s}}}{\text{max}_{\mathbf{s}}} - \tau_{\mathbf{s}} ||_2
\end{equation}
where $\text{min}_{\mathbf{s}} \in \mathbb{R}^3$ and $\text{max}_{\mathbf{s}}\in \mathbb{R}^3$ are the minimum and maximum scales and $\tau_{\mathbf{s}}=0.4$ is the ratio threshold of scales.

\begin{figure*}[!t]
	\newlength\mytmplenn
	\setlength\mytmplenn{.153\linewidth}
	\setlength{\tabcolsep}{1pt}
	\renewcommand{\arraystretch}{0.5}
	\centering
	\begin{tabular}{cccccc}
		Scene & Ground Truth & MANO & HARP & MANUS & Ours \\

		\includegraphics[width=\mytmplenn]{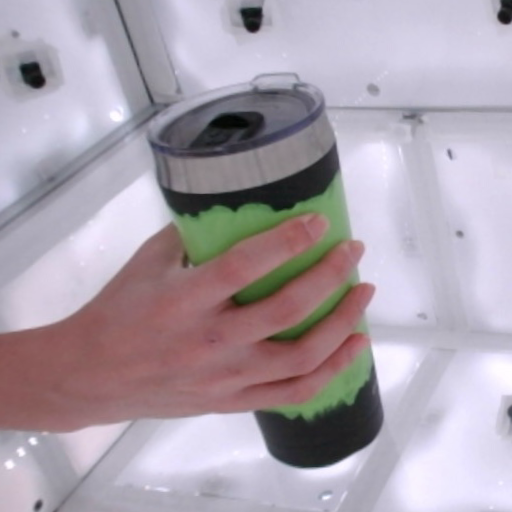} &
		\includegraphics[width=\mytmplenn]{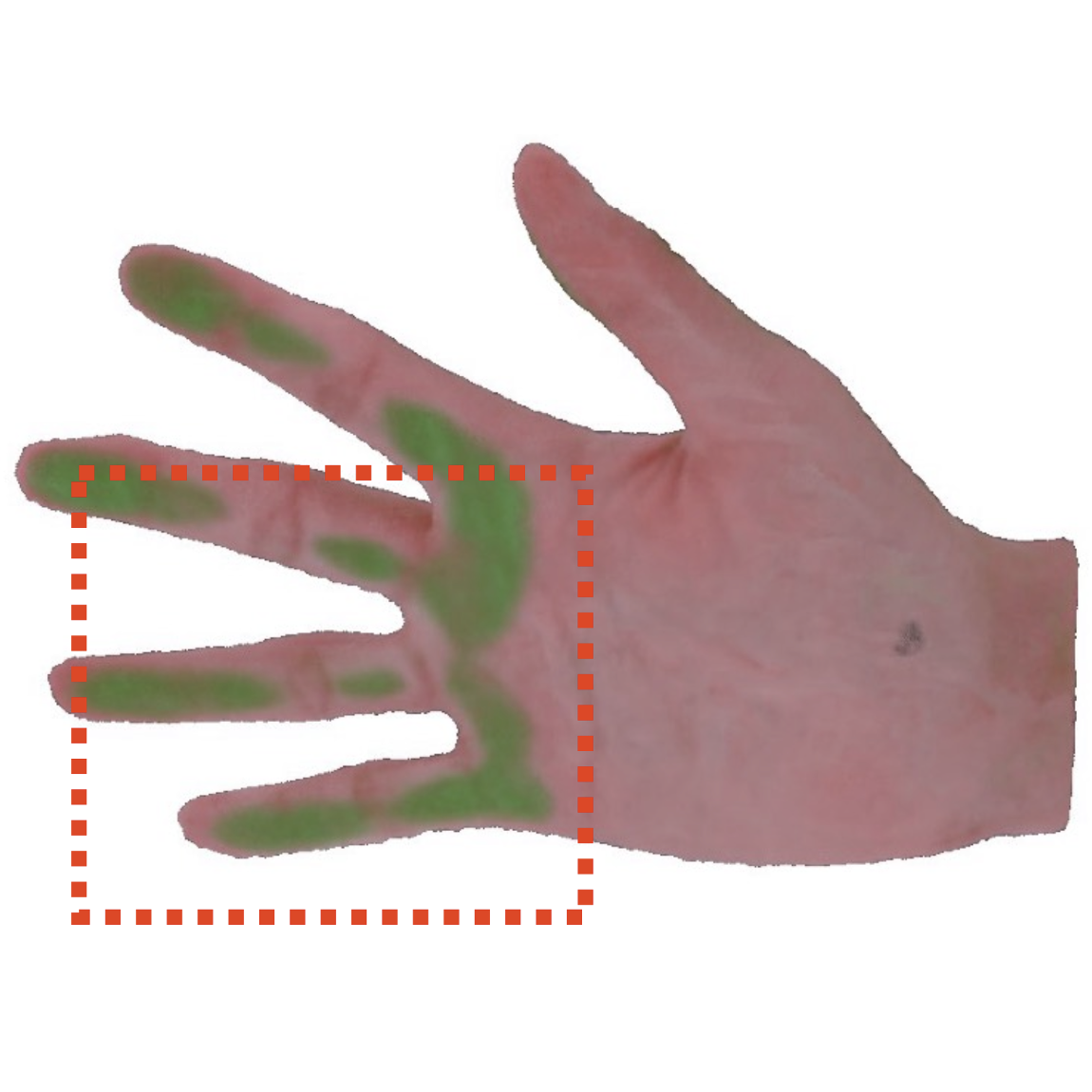} &
		\includegraphics[width=\mytmplenn]{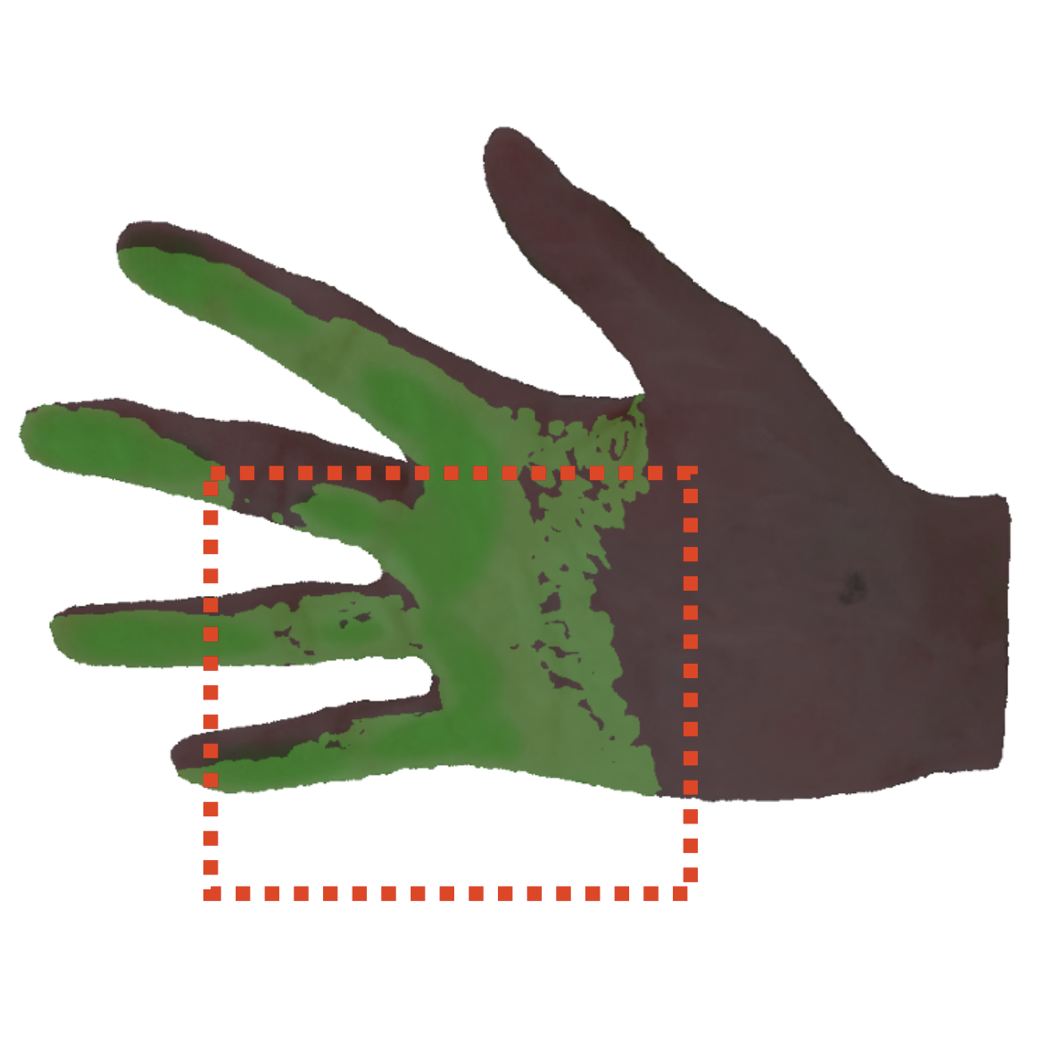} &
		\includegraphics[width=\mytmplenn]{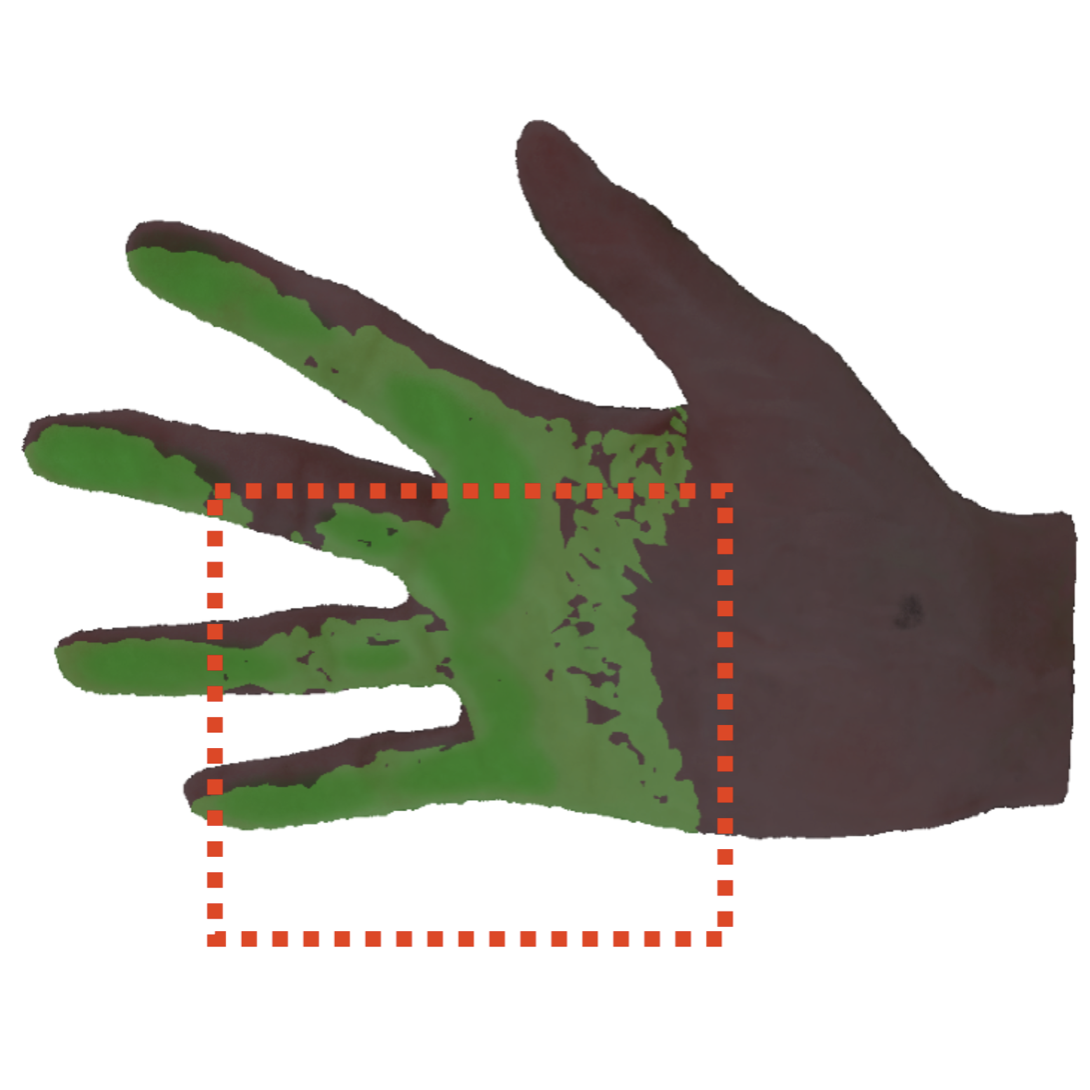} &
		\includegraphics[width=\mytmplenn]{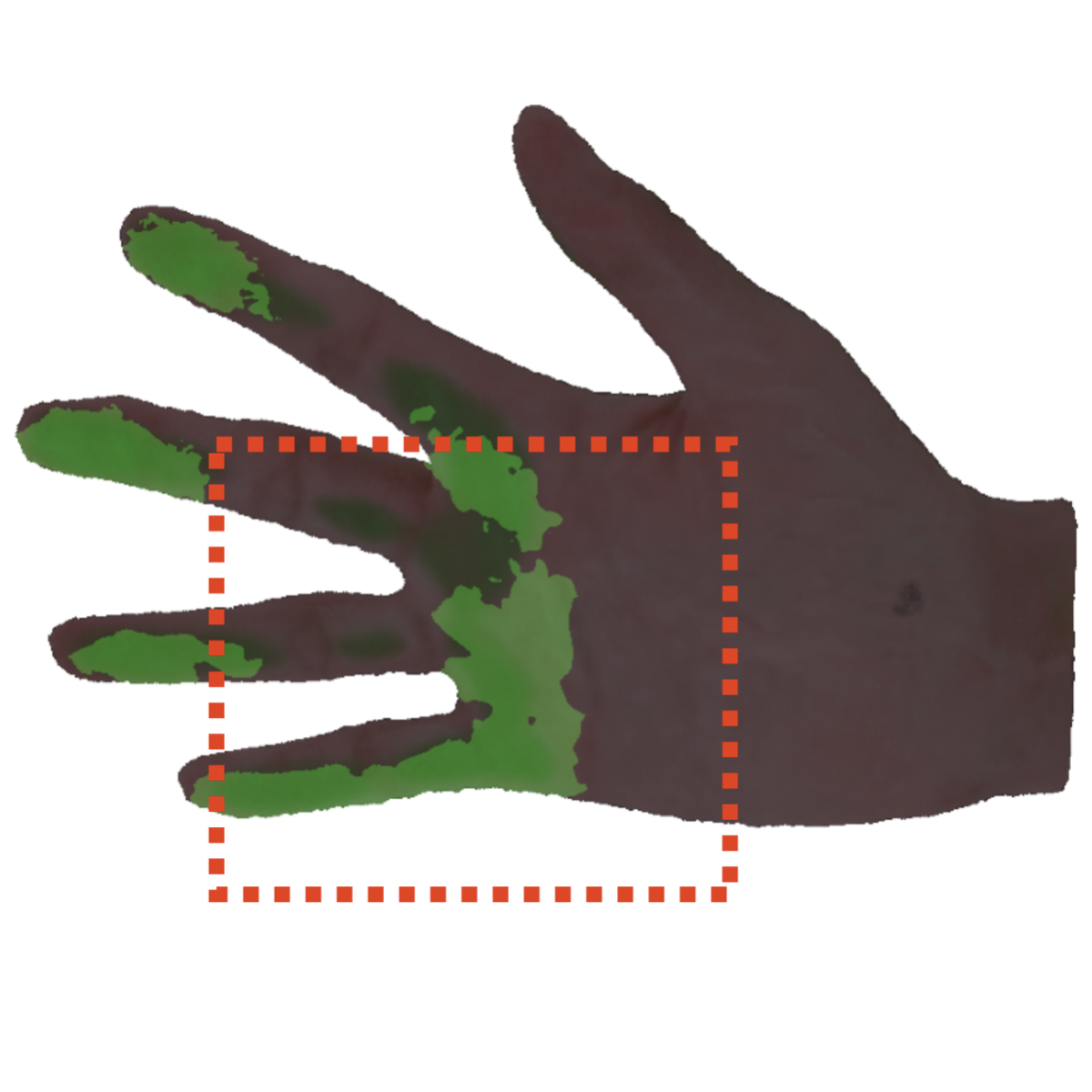} &
            \includegraphics[width=\mytmplenn]{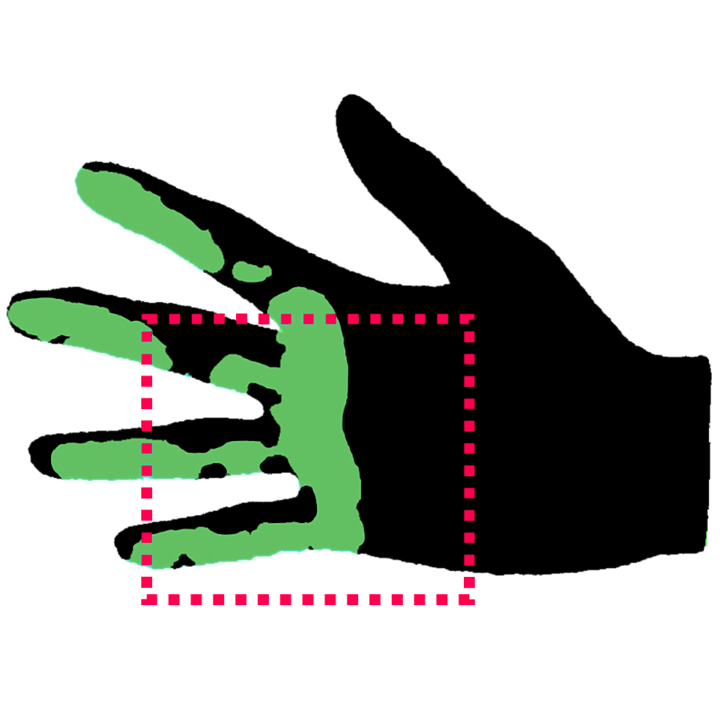} 
		\\
		\includegraphics[width=\mytmplenn]{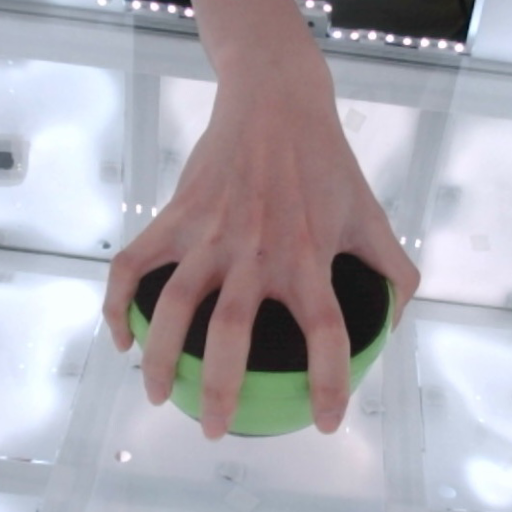} &
		\includegraphics[width=\mytmplenn]{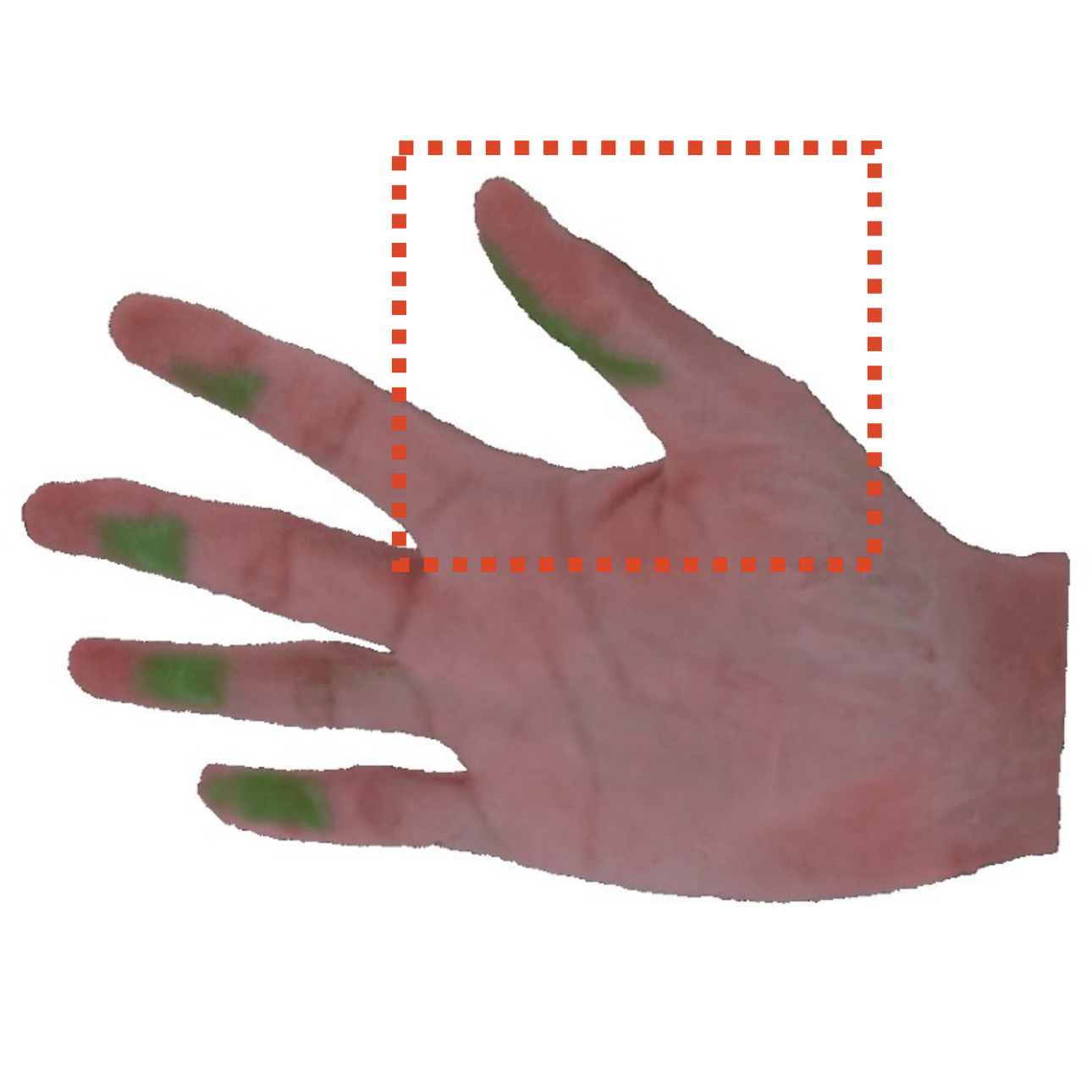} &
		\includegraphics[width=\mytmplenn]{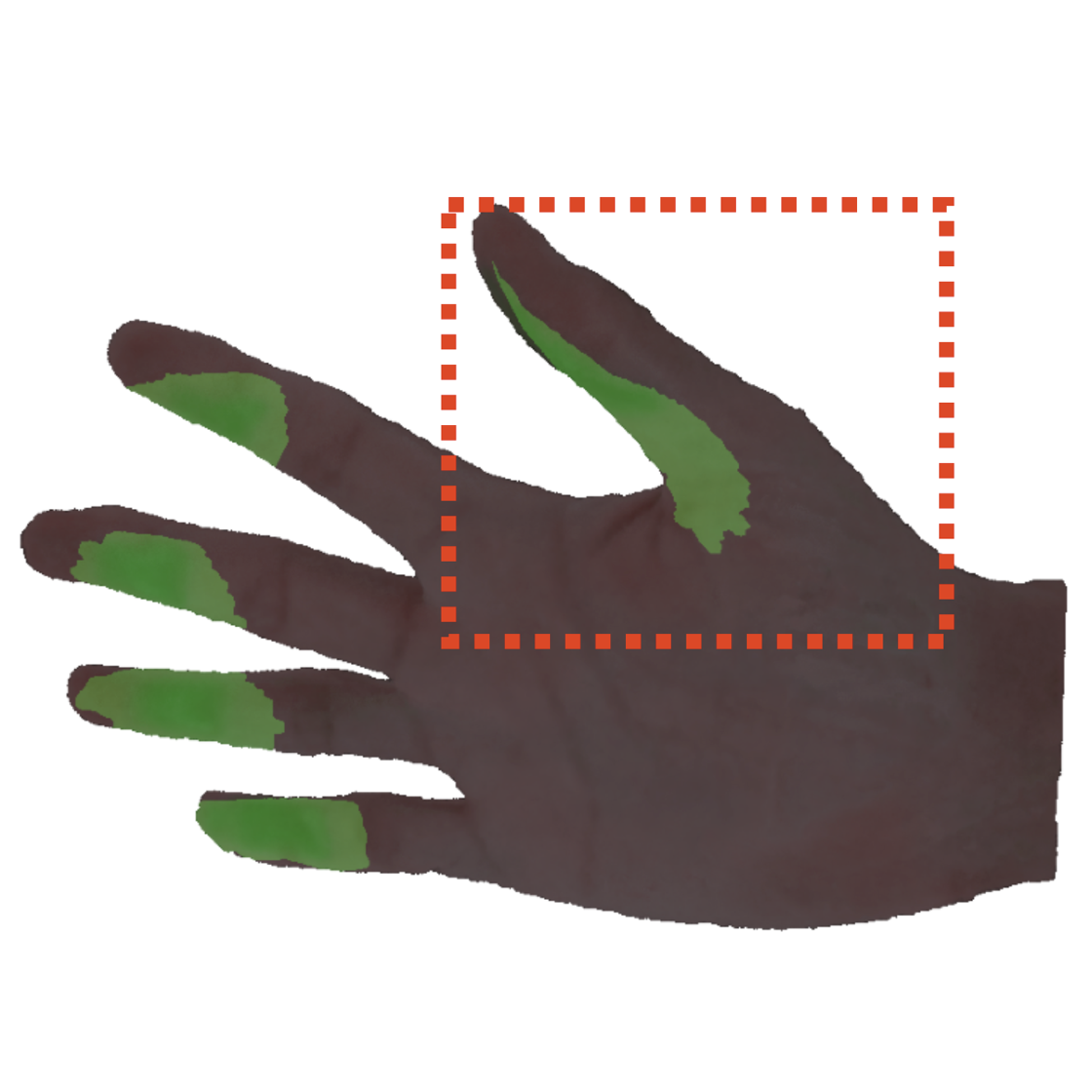} &
		\includegraphics[width=\mytmplenn]{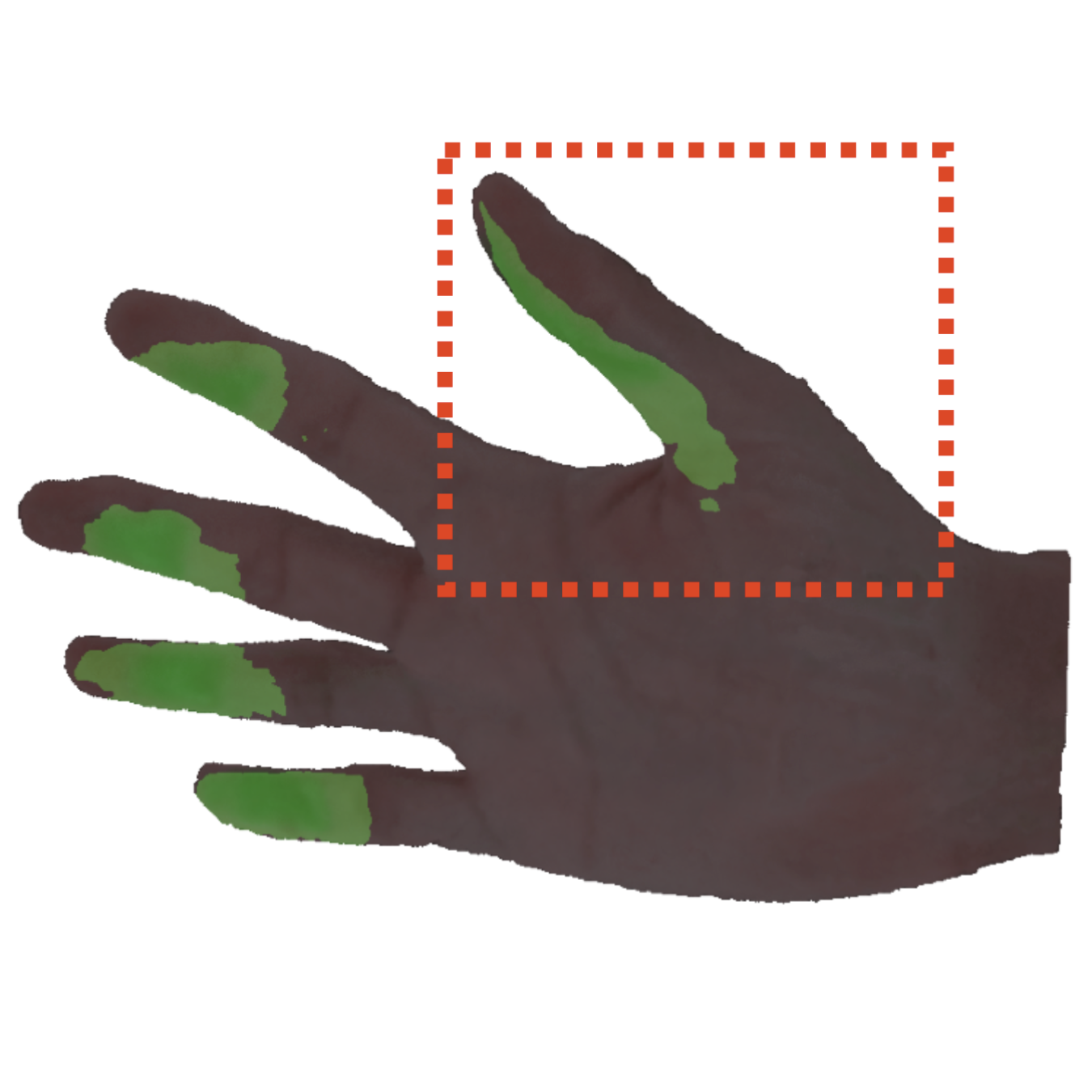} &
		\includegraphics[width=\mytmplenn]{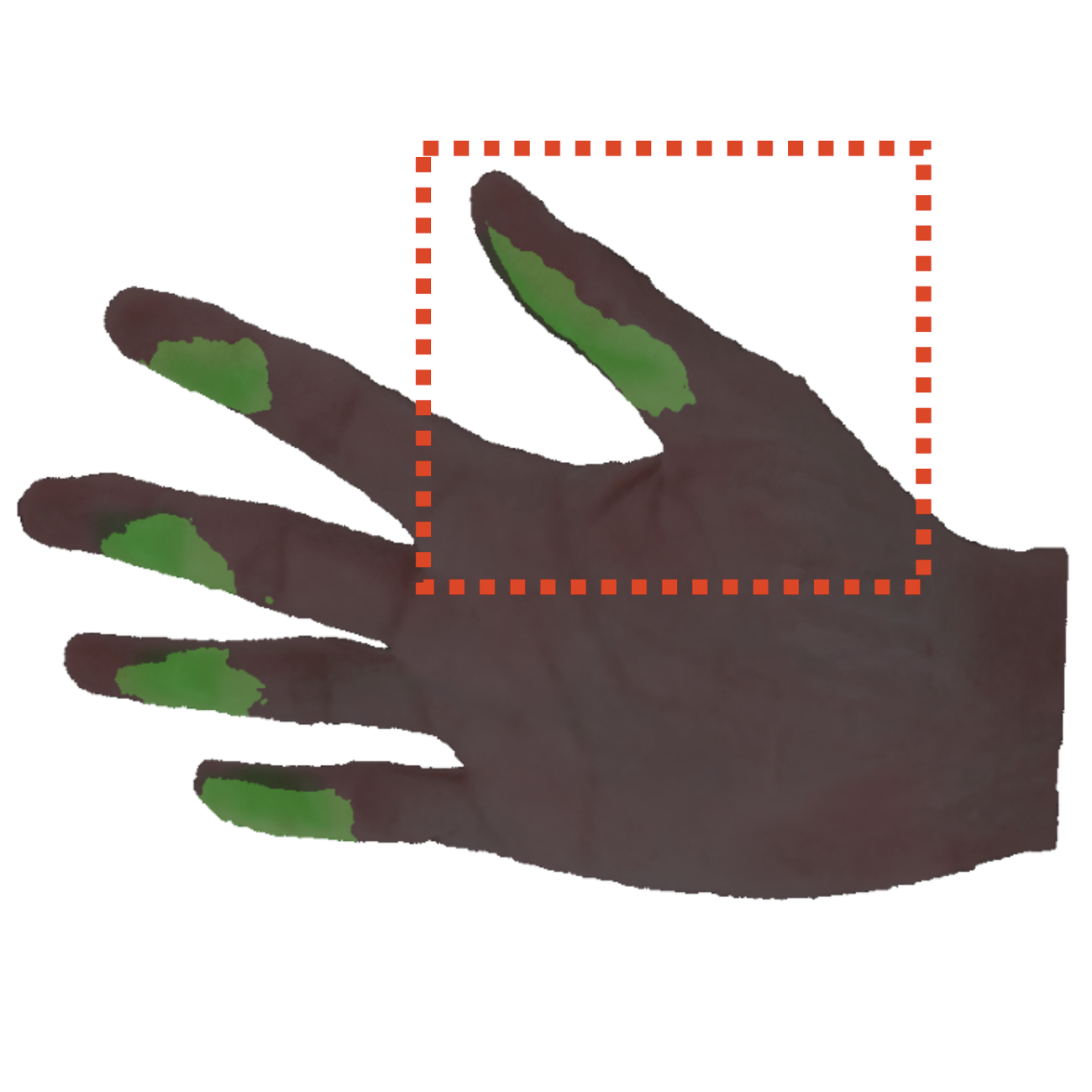} &
            \includegraphics[width=\mytmplenn]{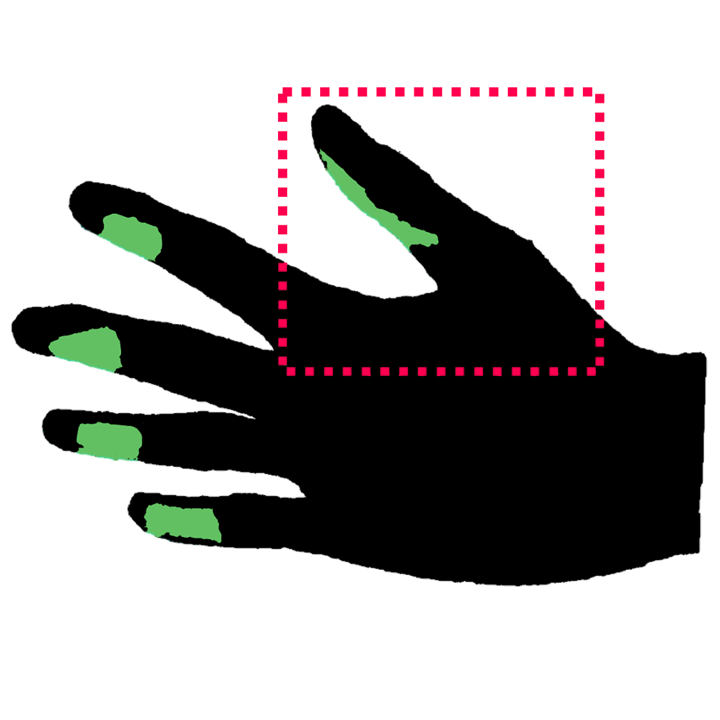} 
		\\
		\includegraphics[width=\mytmplenn]{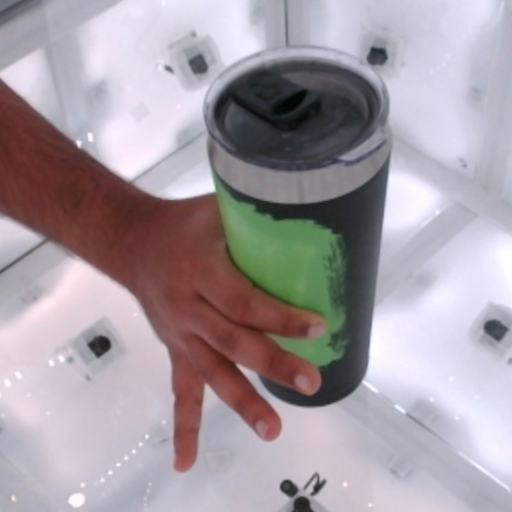} &
		\includegraphics[width=\mytmplenn]{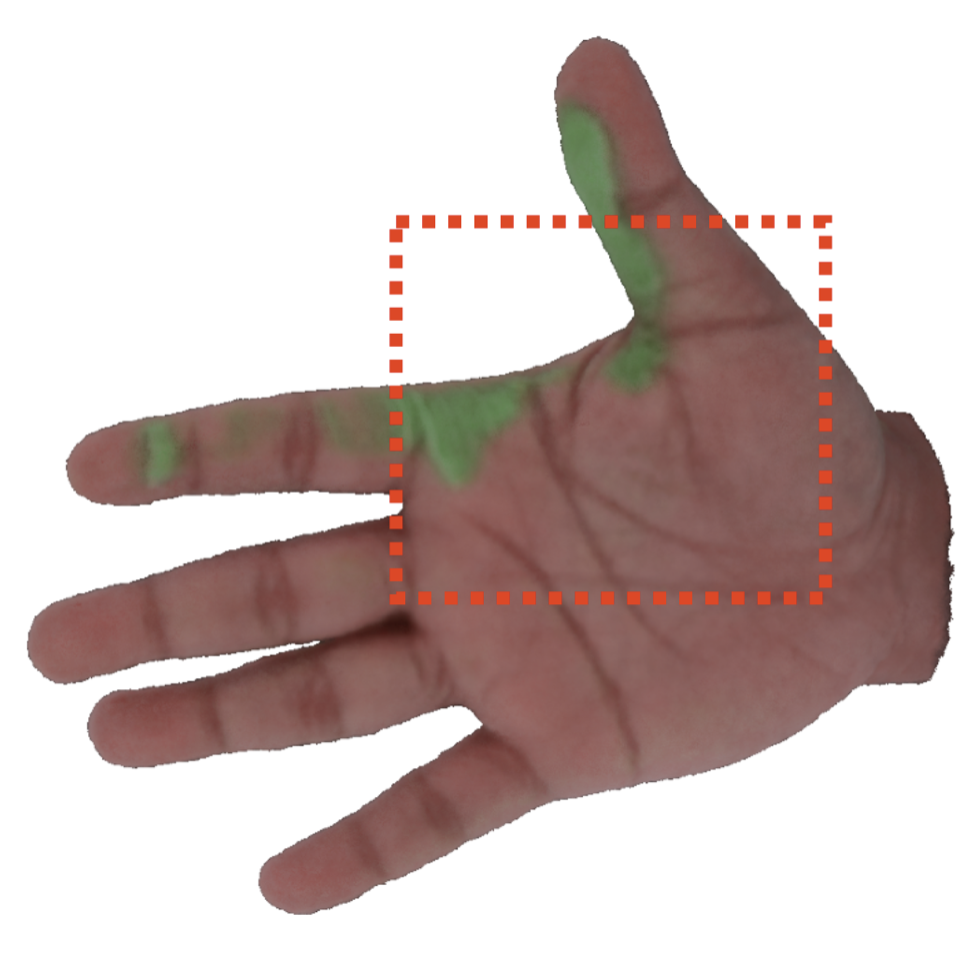} &
		\includegraphics[width=\mytmplenn]{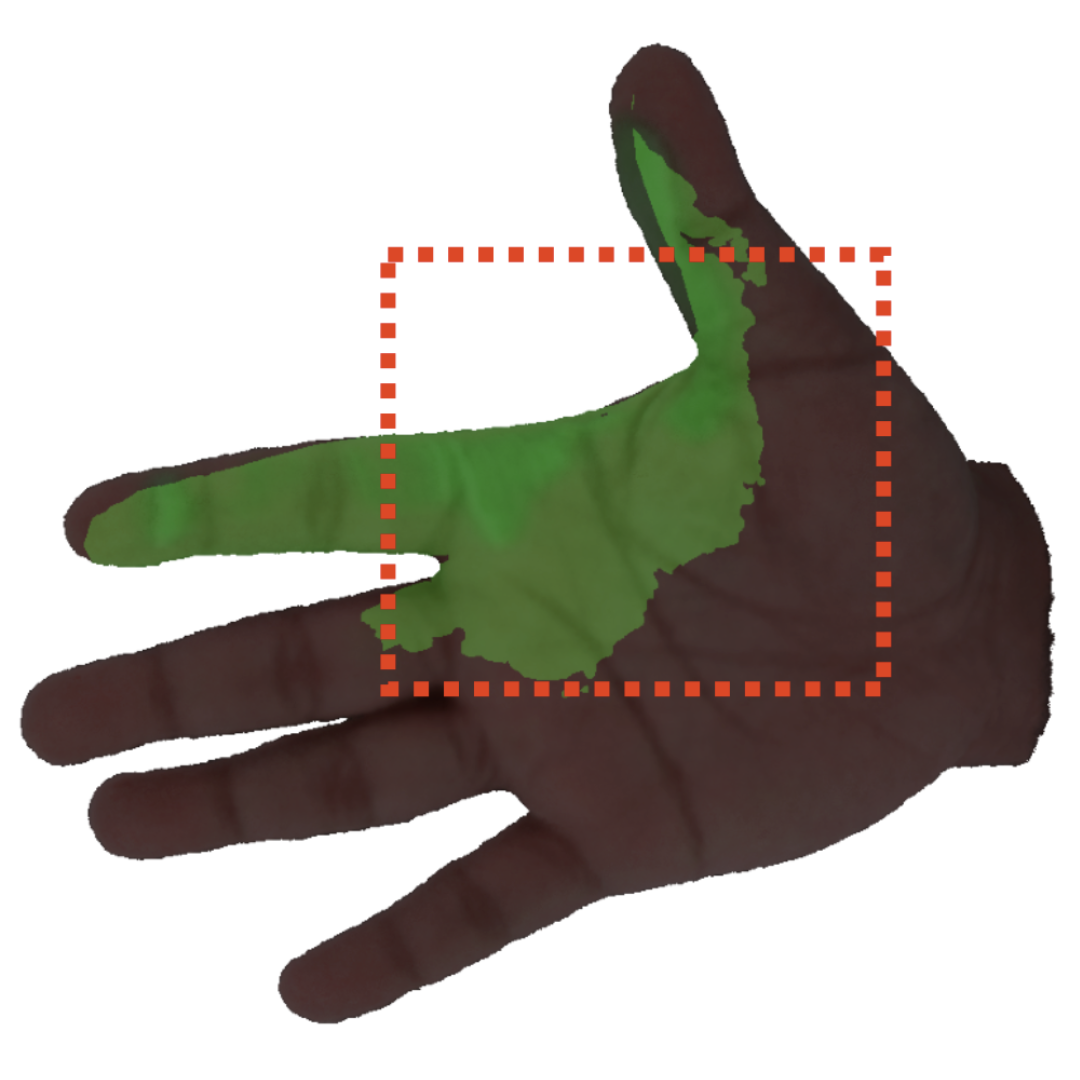} &
		\includegraphics[width=\mytmplenn]{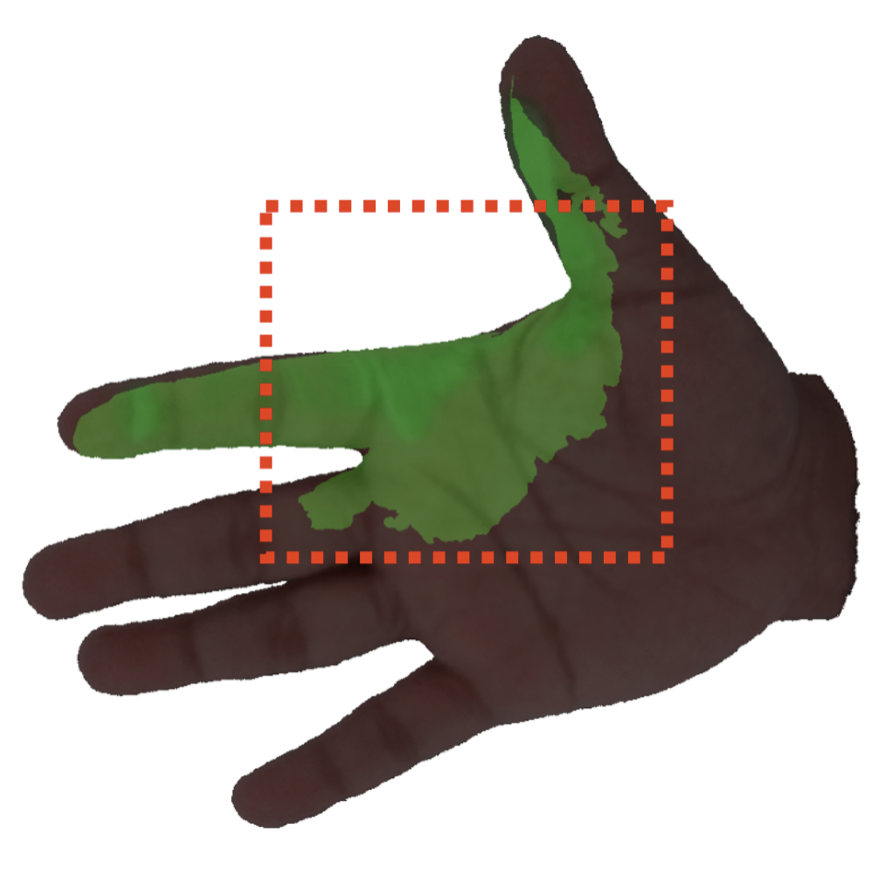} &
		\includegraphics[width=\mytmplenn]{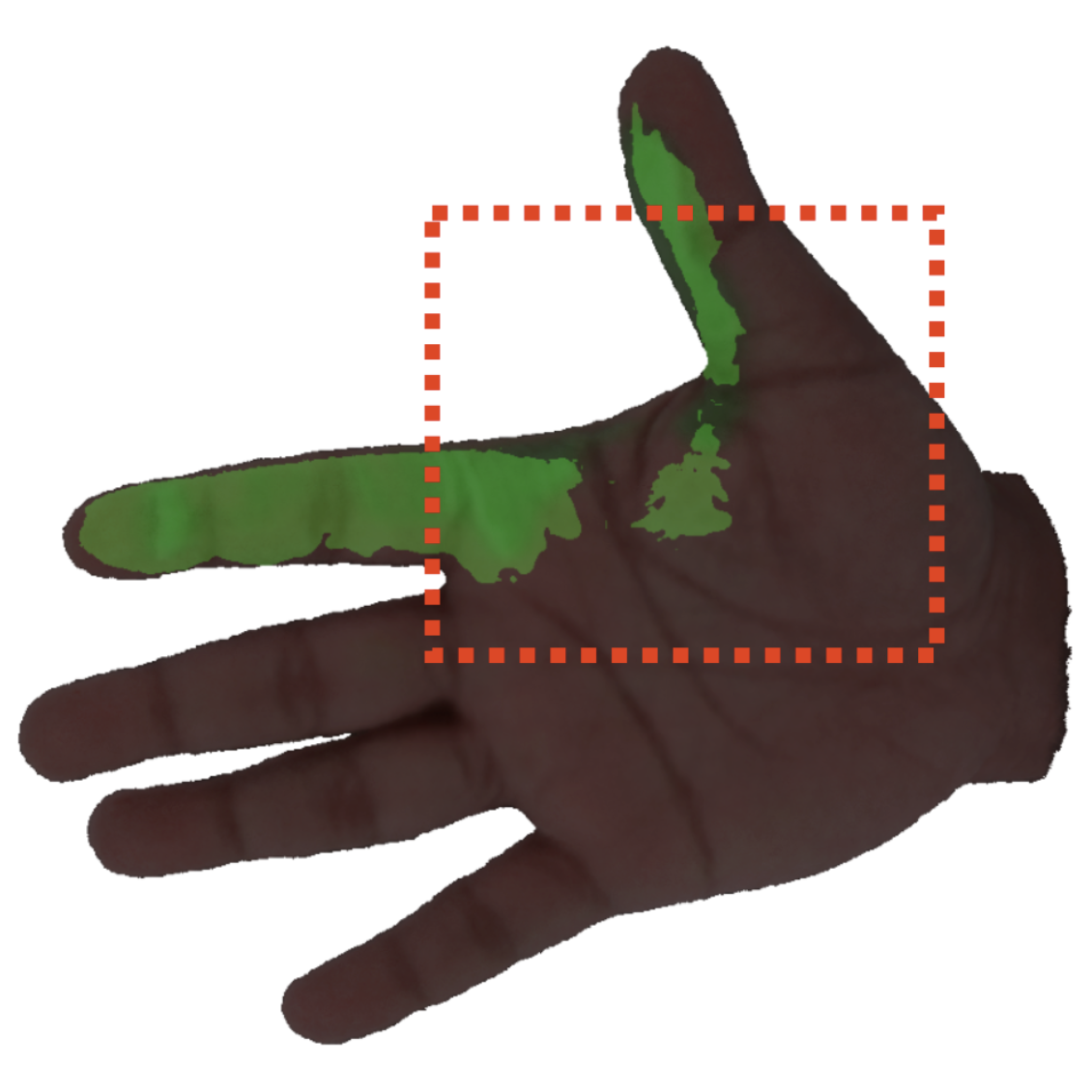} &
            \includegraphics[width=\mytmplenn]{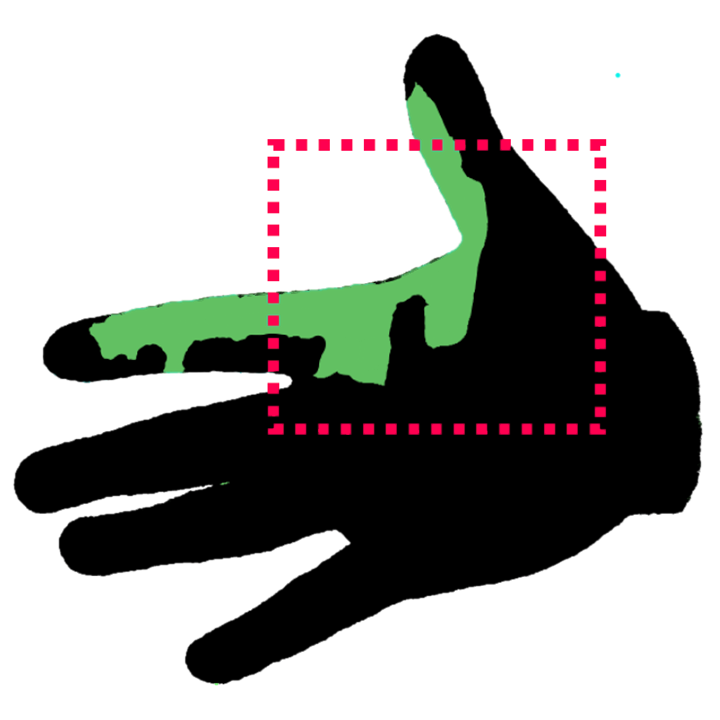} 
		\\
		\includegraphics[width=\mytmplenn]{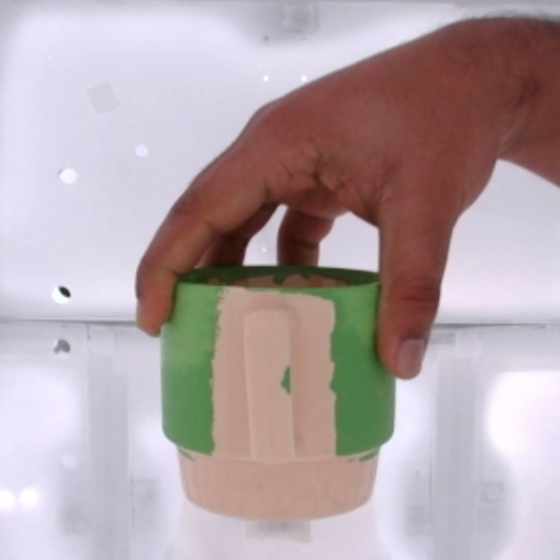} &
		\includegraphics[width=\mytmplenn]{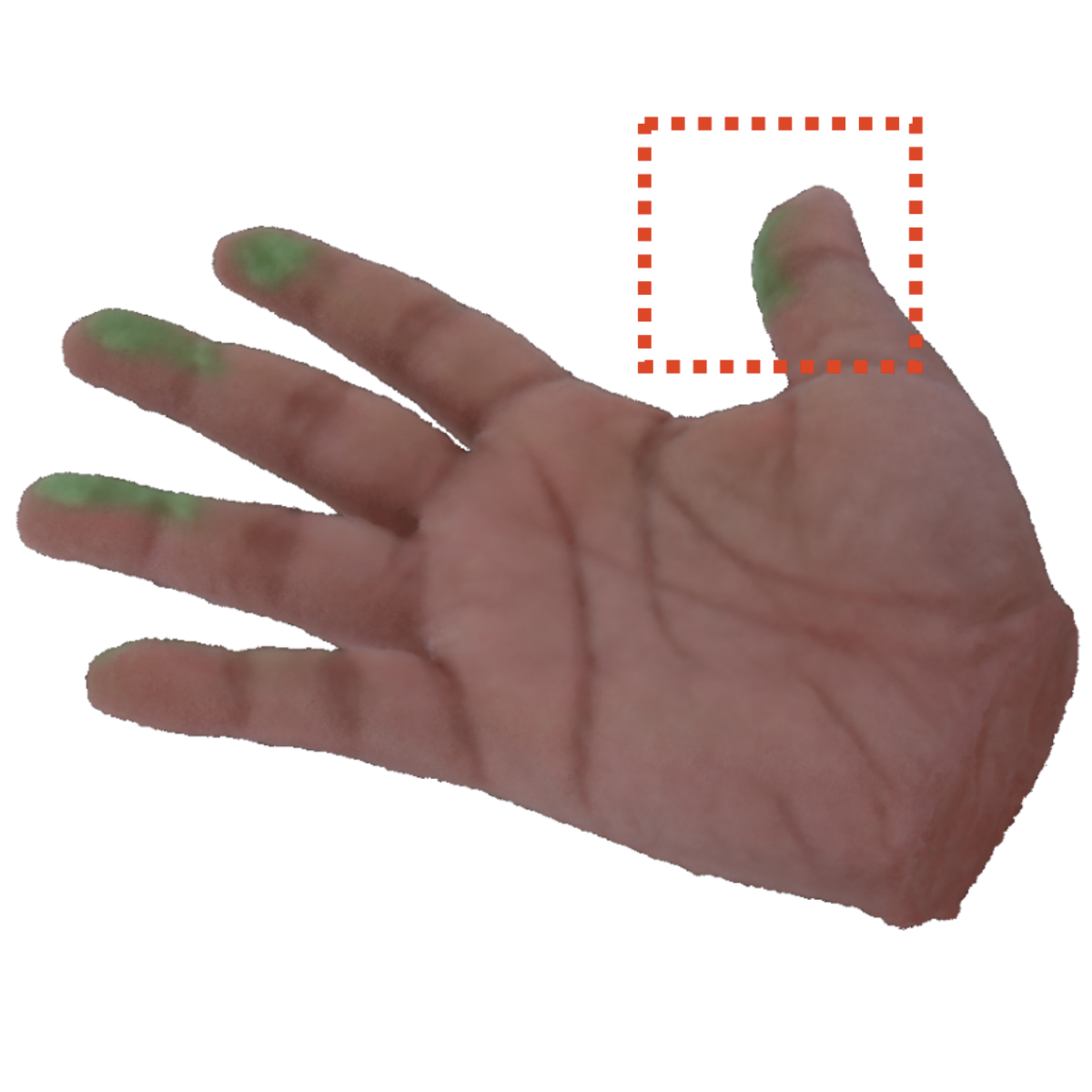} &
		\includegraphics[width=\mytmplenn]{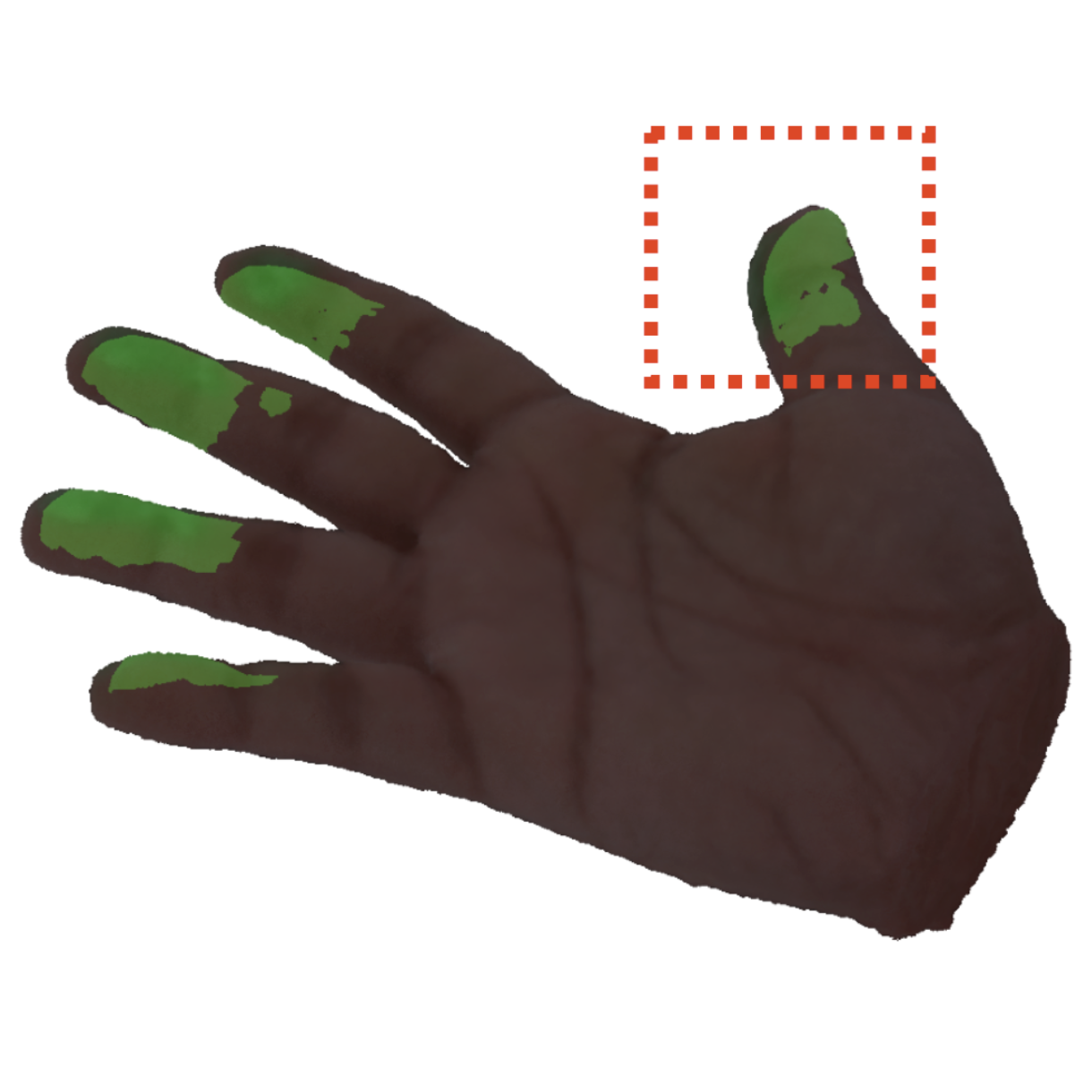} &
		\includegraphics[width=\mytmplenn]{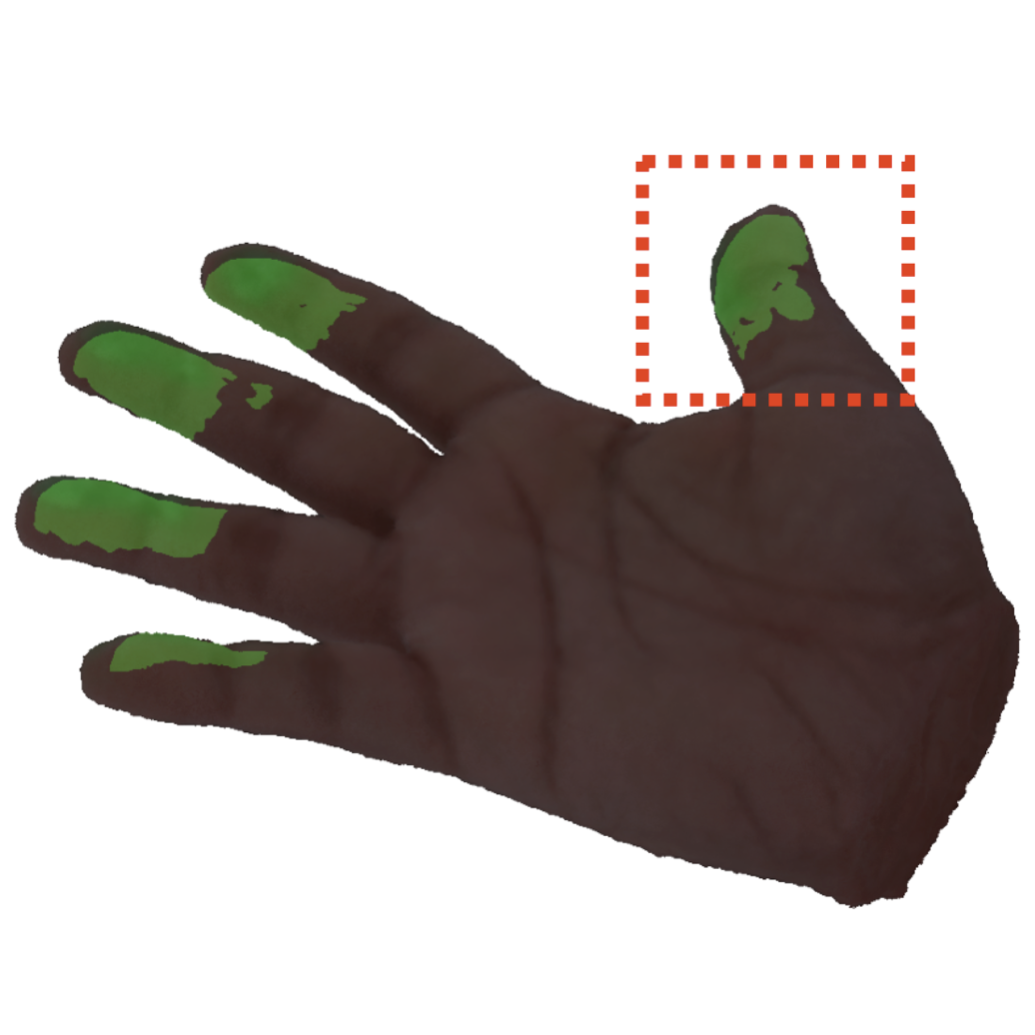} &
		\includegraphics[width=\mytmplenn]{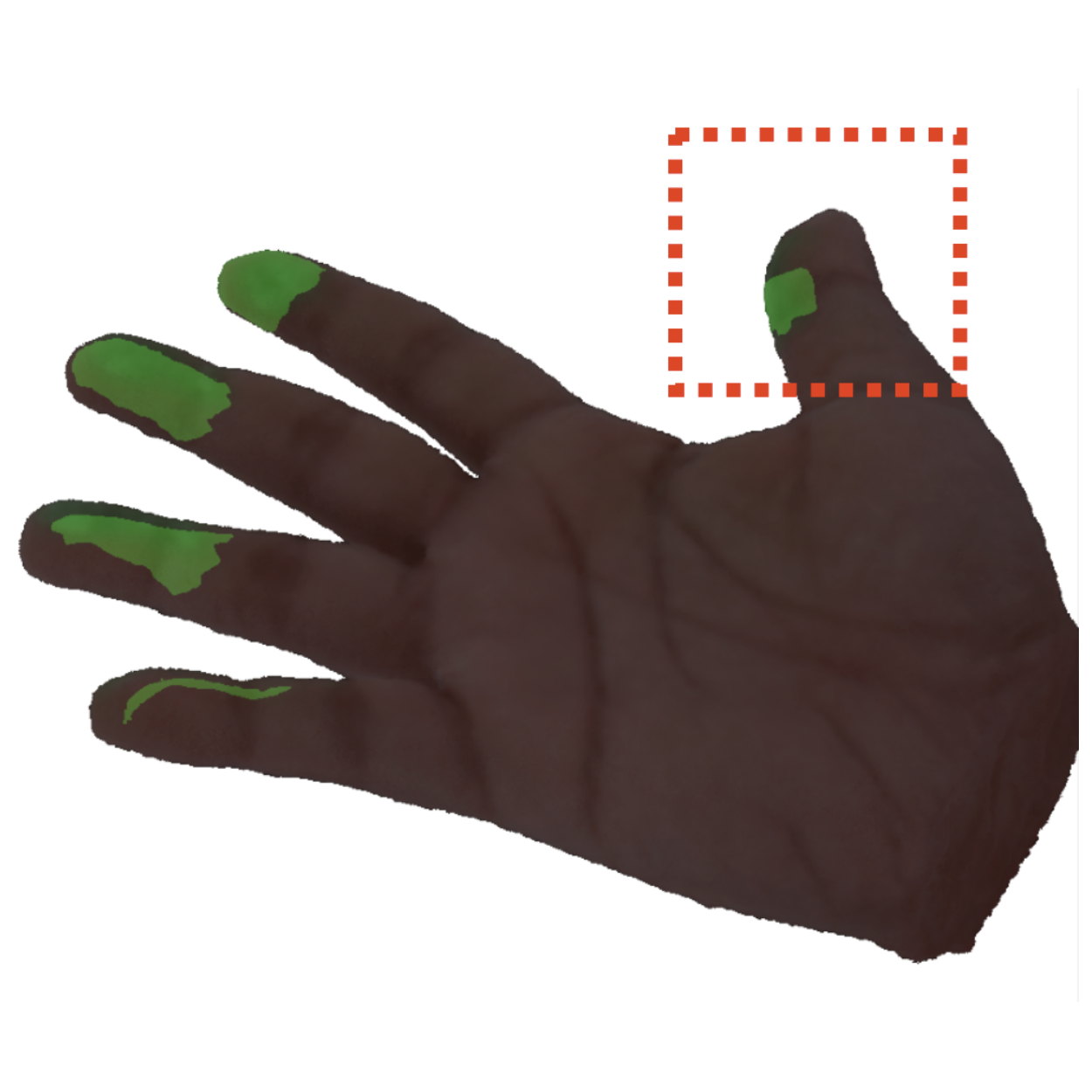} &
            \includegraphics[width=\mytmplenn]{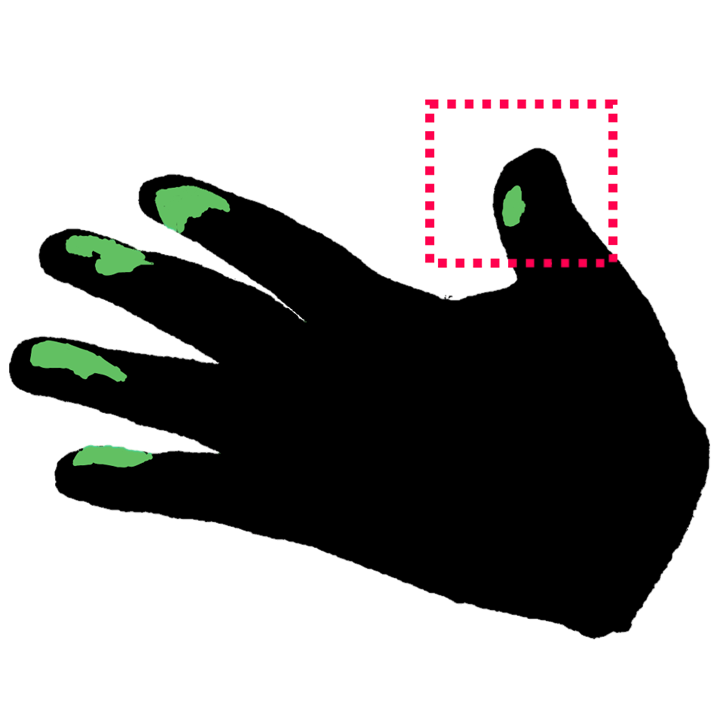} 
		\\
	\end{tabular}
\vspace{-0.05in}
	\caption{
		\label{Fig:contact_compare}
\textbf{Qualitative Comparisons on Contact Estimation.} 
We compare the accumulated contact estimation of \methodname against MANUS~\cite{pokhariya2024manus}, MANO~\cite{MANO:SIGGRAPHASIA:2017}, and HARP~\cite{karunratanakul2023harp} on sequences from \datasetname benchmark. \methodname yields more accurate and coherent accumulated contact estimation that closely aligns with the ground truth, contrasting sharply with the over-segmentation and noisy artifacts observed in baselines.
}
\vspace{-0.1in}
\end{figure*}

\paragraph{Overall Loss Terms.}
We supervise the rendered images by photometric loss $\mathcal{L}_{C}$ combining $\mathcal{L}_1$ with a D-SSIM term~\cite{kerbl3Dgaussians, Huang2DGS2024} $\mathcal{L}_{\text{D\_SSIM}}$:
\begin{equation}
    \mathcal{L}_c = (1-\lambda)\mathcal{L}_1 + \lambda \mathcal{L}_{\text{D\_SSIM}}
\end{equation}
where $\lambda=0.2$. 
Following 2DGS~\cite{Huang2DGS2024}, we use the depth distortion term $\mathcal{L}_d$ that encourages the concentration of the 2D Gaussian surfels by adjusting the intersection depth and the normal consistency loss $\mathcal{L}_n$ that encourages 2D Gaussian surfels locally approximate the surface by aligning their normals with the estimated surface normals. Following GaussianAvatars~\cite{qian2024gaussianavatarsphotorealisticheadavatars}, we use two rigging regularization terms $\mathcal{L}_p$ and $\mathcal{L}_s$ to restrict the position and scale of hands' Gaussian surfels for better alignment with their parent triangles. The overall loss function is:
\begin{equation}
    \mathcal{L} = \mathcal{L}_c + \lambda_1 \mathcal{L}_d + \lambda_2 \mathcal{L}_n + \lambda_3 \mathcal{L}_p + \lambda_4 \mathcal{L}_s + \lambda_5 \mathcal{L}_i,
\end{equation}
where $\lambda_1$, $\lambda_2$, $\lambda_3$, $\lambda_4$, and $\lambda_5$ are $100, 0.005, 0.01, 1$ and $0.1$ respectively.

%% file: sec/5_experiment.tex
\begin{table}[t]
\small
\vspace{-0.1in}
\scalebox{1}{
    \begin{tabular}{p{1.3cm} |
>{\centering\arraybackslash}p{1.15cm} >{\centering\arraybackslash}p{1.15cm} 
>{\centering\arraybackslash}p{1.15cm} >{\centering\arraybackslash}p{1.15cm} 
}

         & MANO  & HARP & MANUS & Ours\\
        \hline 
        mIoU$^\uparrow$ & 0.168 & \cellcolor{yellow!40}0.182 & \cellcolor{orange!40}0.211 & \cellcolor{red!40}0.226 \\
        $\text{F}1$ score$^\uparrow$ & 0.279 & \cellcolor{yellow!40}0.299 & \cellcolor{orange!40}0.343 & \cellcolor{red!40}0.378  \\ 
        
    \end{tabular}
}
\vspace{-0.05in}
\caption{
    \label{TAB:contact_comparison} {\textbf{Quantitative Comparisons on the Accumulated Contacts Estimation.} We evaluate the accuracy of the accumulated contacts estimation against other methods and demonstrate consistent improvements across all metrics.}
}
\vspace{-0.1in}
\end{table}

\section{Experiments}
\label{Sec:Experiments}

\subsection{\datasetname Dynamic Contact Benchmark}
Evaluating dynamic hand-object contacts remains a significant challenge due to the difficulty of obtaining reliable ground truth data.
Existing real-world hand–object manipulation datasets~\cite{kwon2021h2o, ohkawa2023assemblyhands, liu2022hoi4d, fan2023arctic, banerjee2024introducing, zhan2024oakink2, fu2024gigahandsmassiveannotateddataset, lu2024diva, liu2024taco, taheri2020grab} lack ground-truth contact annotations, limiting their utility for quantitative evaluation.
To address this gap, we introduce \datasetname, a new benchmark featuring diverse real-world hand–object manipulation sequences with accurate ground-truth accumulated contacts.
\datasetname includes $15$ single-hand grasping sequences from MANUS-Grasps~\cite{pokhariya2024manus}, along with 6 newly captured complex bimanual manipulation sequences.

These additional sequences were collected using an enhanced version of the wet-paint residue method~\cite{kamakura1980patterns, pokhariya2024manus}, redesigned to support scalable capture of complex interactions.
We defer the details of our capture procedure to the supplementary material.


In total, \datasetname provides $21$ real-world tabletop manipulation sequences with ground-truth accumulated contact, benchmarking dynamic contact estimation methods.

\begin{figure*}[!h]
	\newlength\mytmplen
	\setlength\mytmplen{.14\linewidth}
	\setlength{\tabcolsep}{1pt}
	\renewcommand{\arraystretch}{0.2}
	\centering
        \small
	\begin{tabular}{ccccccc}
		Ground Truth & Ours & Deformable3DGS   & 4DGaussians & Realtime4DGS & AT-GS & 3DGStream \\

            \includegraphics[width=\mytmplen,trim={3pt 3pt 3pt 3pt},clip]{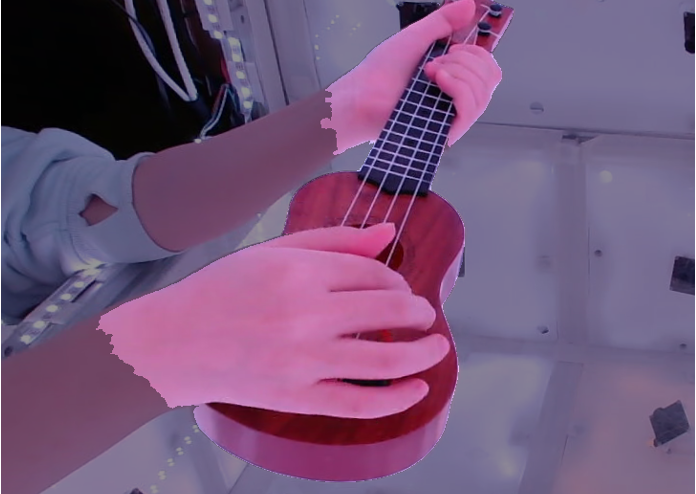} &
		\includegraphics[width=\mytmplen,trim={3pt 3pt 3pt 3pt},clip]{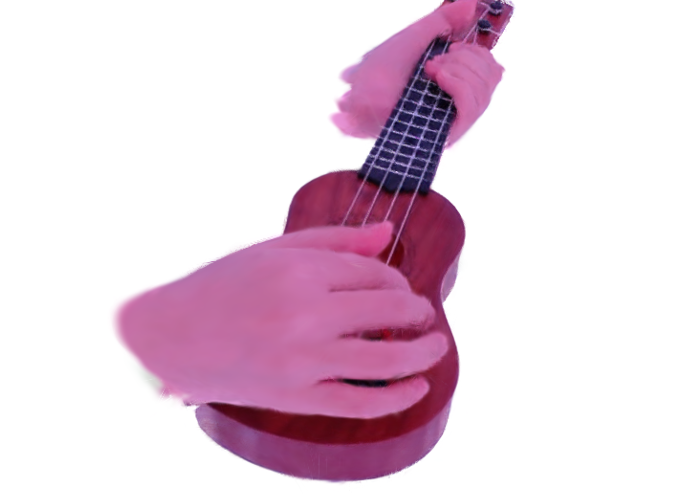} &
		\includegraphics[width=\mytmplen,trim={3pt 3pt 3pt 3pt},clip]{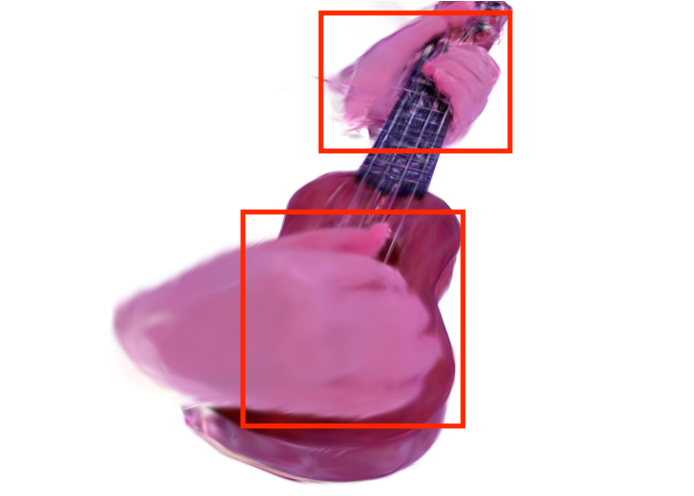} &
		\includegraphics[width=\mytmplen,trim={3pt 3pt 3pt 3pt},clip]{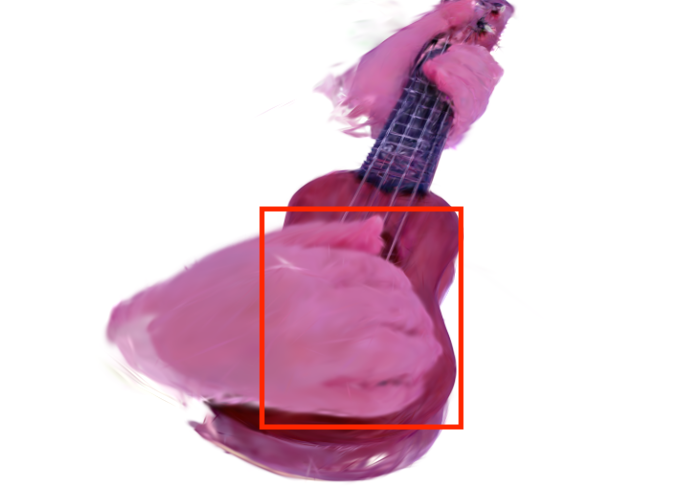} &
		\includegraphics[width=\mytmplen,trim={3pt 3pt 3pt 3pt},clip]{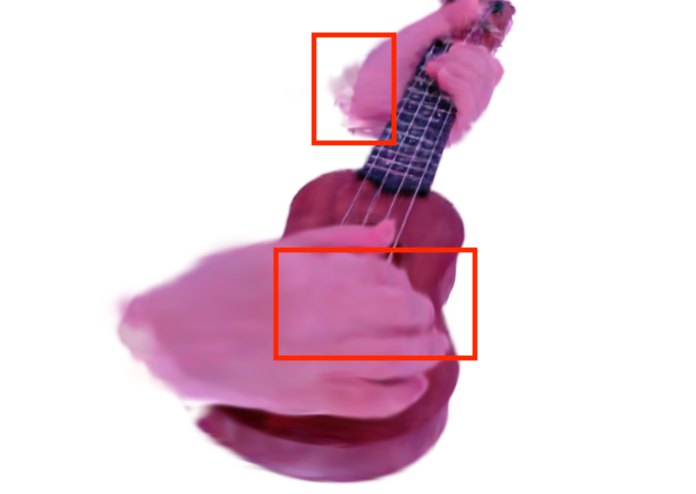} &
            \includegraphics[width=\mytmplen,trim={3pt 3pt 3pt 3pt},clip]{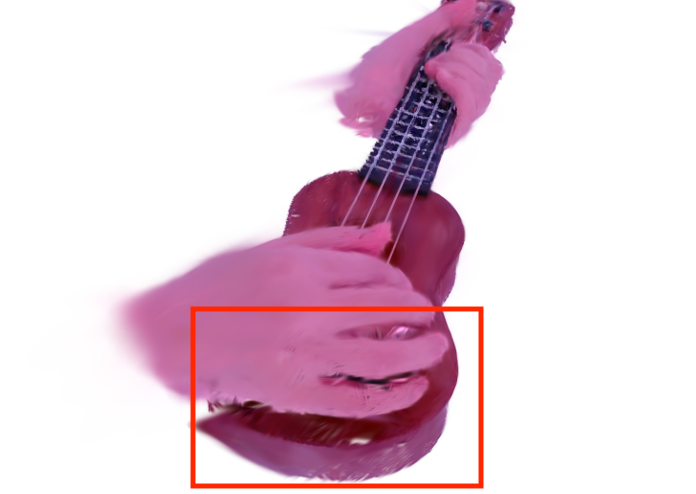}  &
            \includegraphics[width=\mytmplen,trim={3pt 3pt 3pt 3pt},clip]{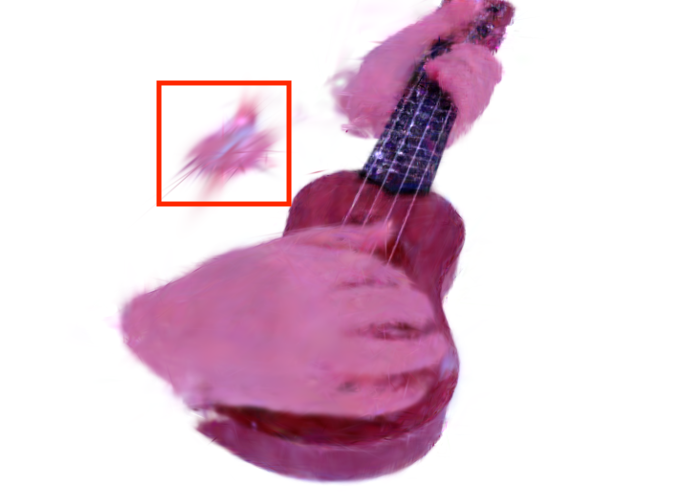}
            \\
		\includegraphics[width=\mytmplen,trim={3pt 3pt 3pt 3pt},clip]{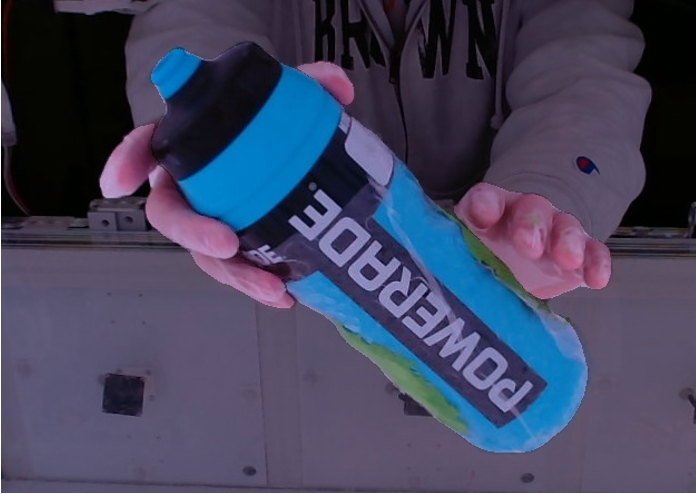} &
		\includegraphics[width=\mytmplen,trim={3pt 3pt 3pt 3pt},clip]{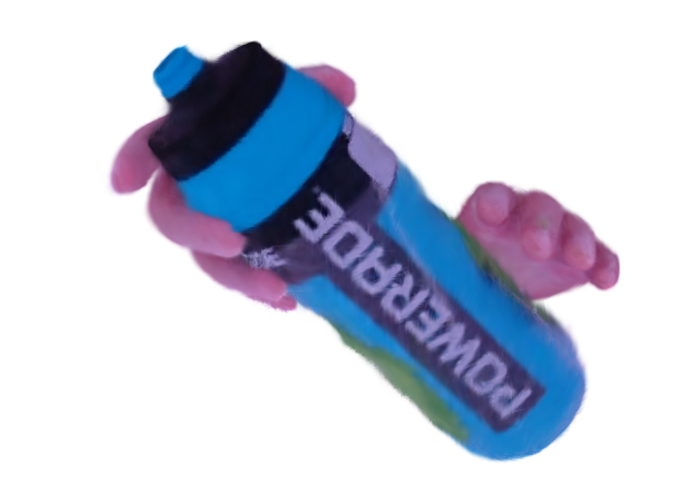} &
		\includegraphics[width=\mytmplen,trim={3pt 3pt 3pt 3pt},clip]{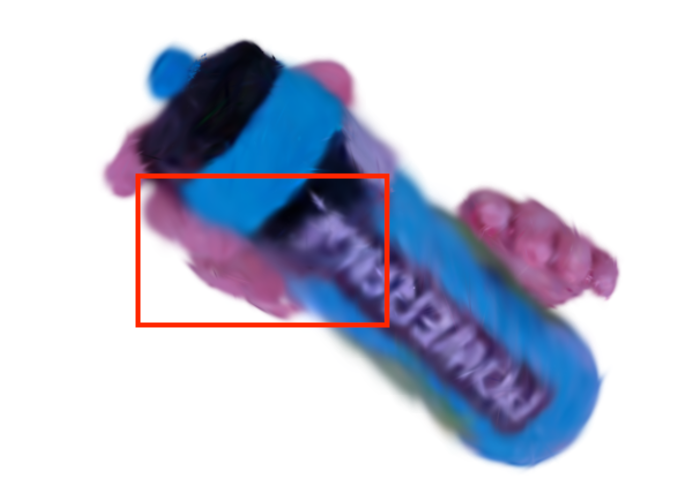} &
		\includegraphics[width=\mytmplen,trim={3pt 3pt 3pt 3pt},clip]{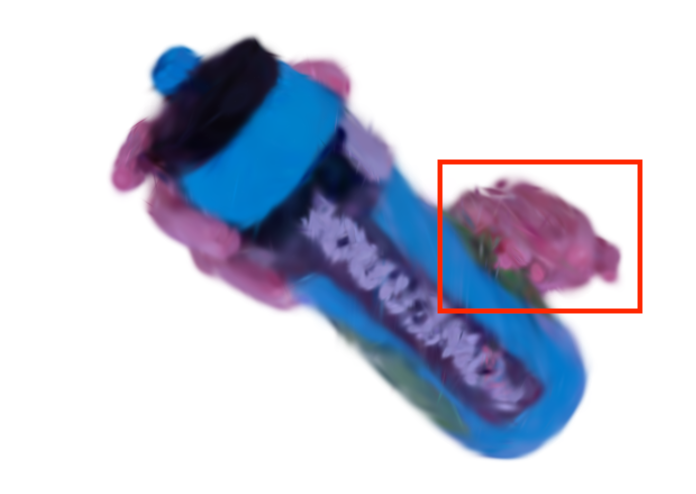} &
		\includegraphics[width=\mytmplen,trim={3pt 3pt 3pt 3pt},clip]{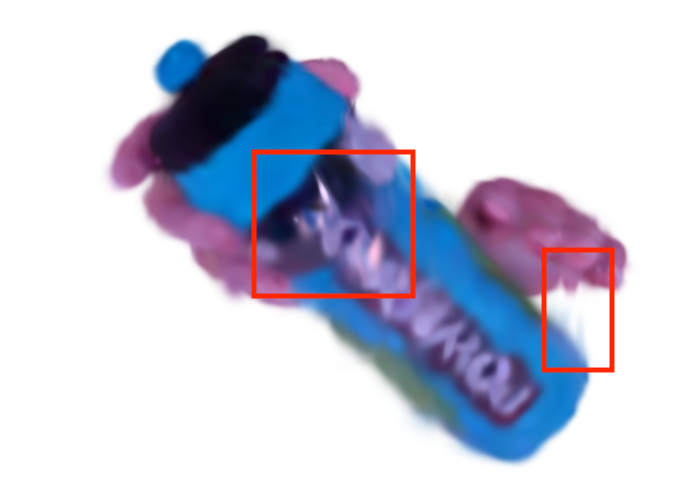} &
            \includegraphics[width=\mytmplen,trim={3pt 3pt 3pt 3pt},clip]{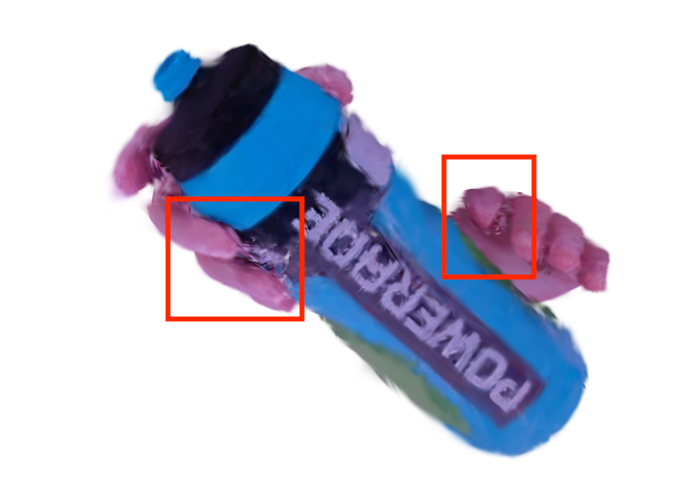}  &
            \includegraphics[width=\mytmplen,trim={3pt 3pt 3pt 3pt},clip]{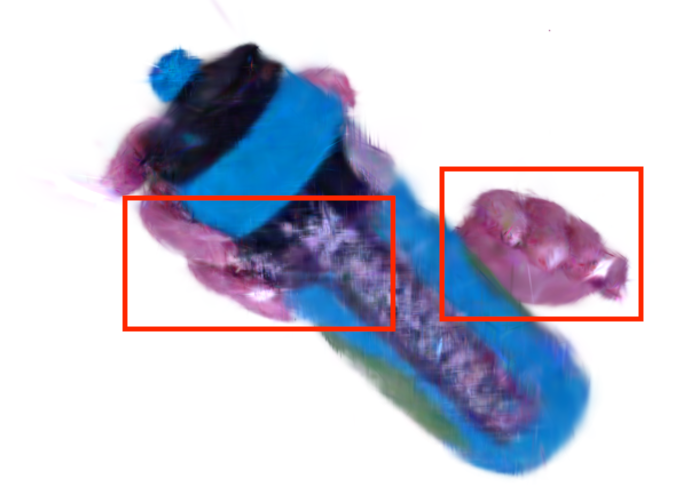}
            \\
		\includegraphics[width=\mytmplen]{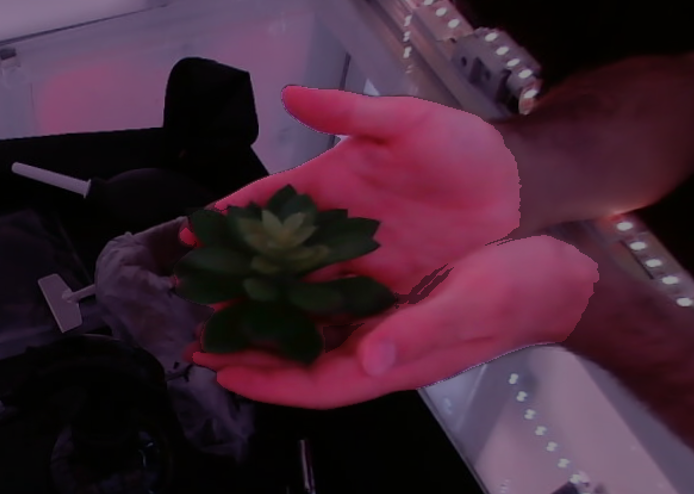} &
		\includegraphics[width=\mytmplen]{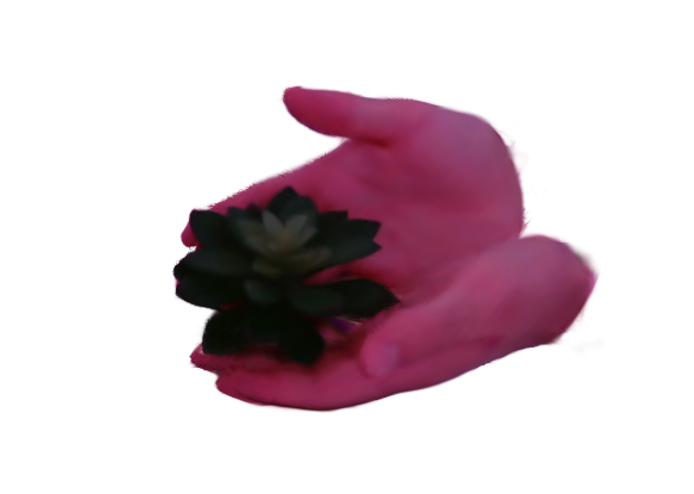} &
		\includegraphics[width=\mytmplen]{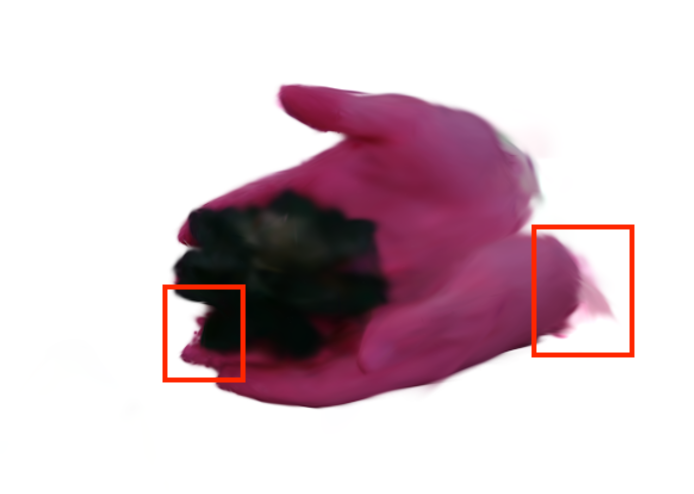} &
		\includegraphics[width=\mytmplen]{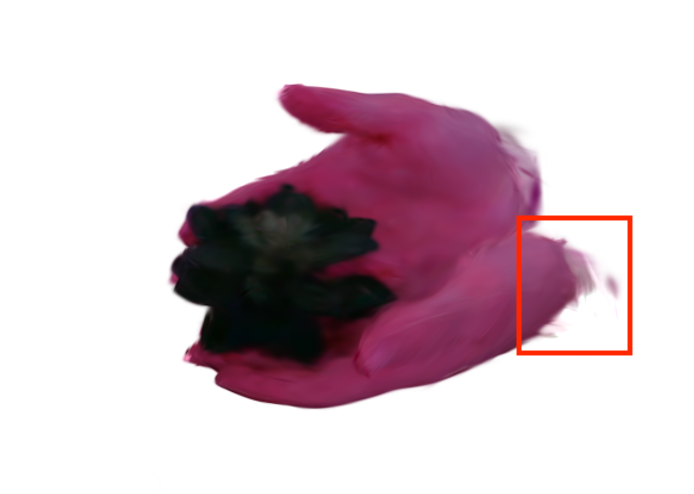} &
		\includegraphics[width=\mytmplen]{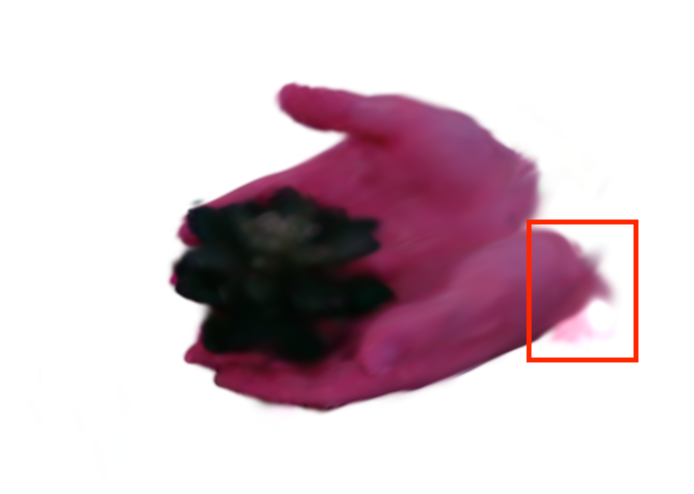} &
            \includegraphics[width=\mytmplen]{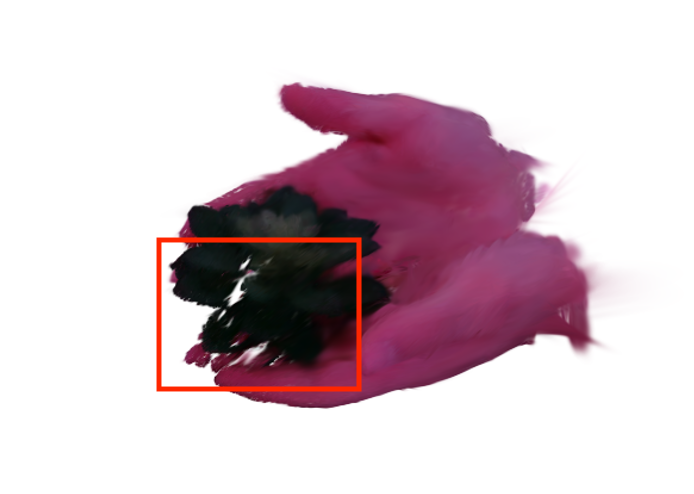}  &
            \includegraphics[width=\mytmplen]{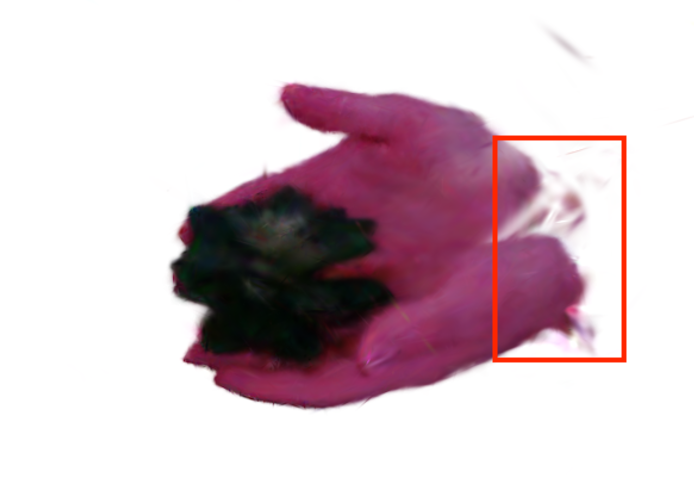}
		\\
		\includegraphics[width=\mytmplen,trim={3pt 3pt 3pt 3pt},clip]{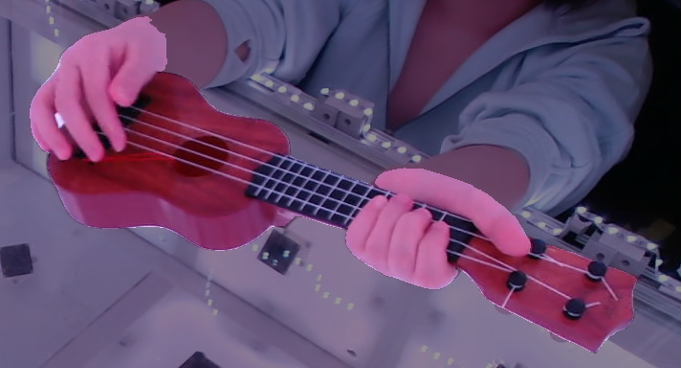} &
		\includegraphics[width=\mytmplen,trim={3pt 3pt 3pt 3pt},clip]{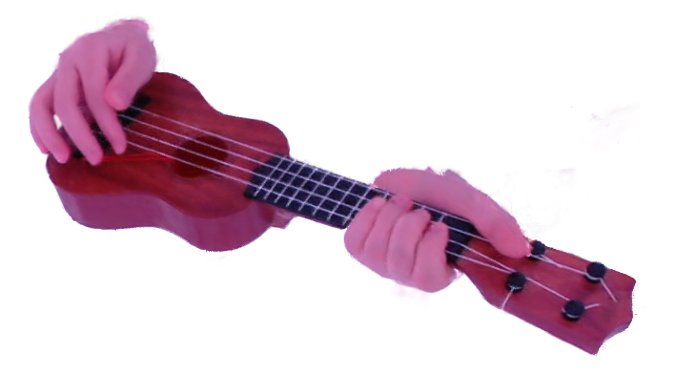} &
		\includegraphics[width=\mytmplen,trim={3pt 3pt 3pt 3pt},clip]{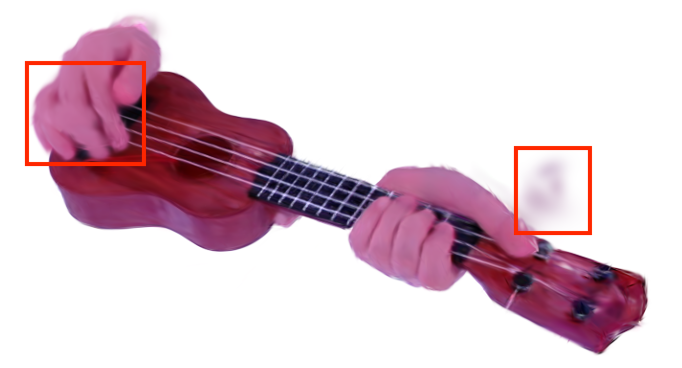} &
		\includegraphics[width=\mytmplen,trim={3pt 3pt 3pt 3pt},clip]{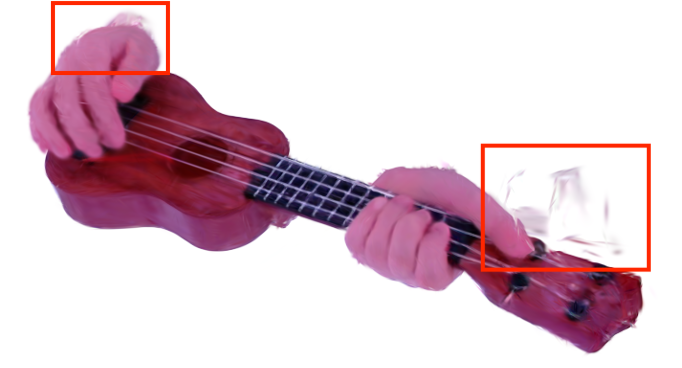} &
		\includegraphics[width=\mytmplen,trim={3pt 3pt 3pt 3pt},clip]{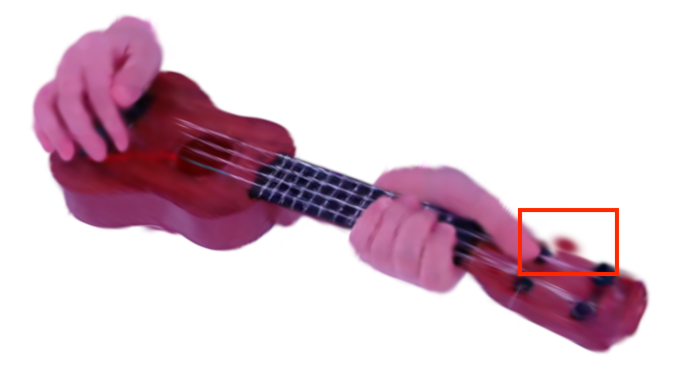} &
            \includegraphics[width=\mytmplen,trim={3pt 3pt 3pt 3pt},clip]{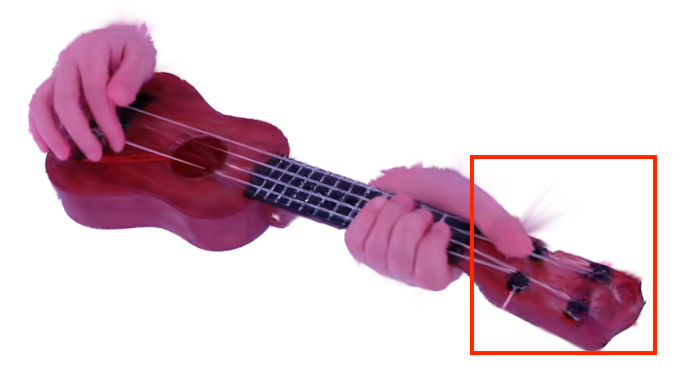}  &
            \includegraphics[width=\mytmplen,trim={3pt 3pt 3pt 3pt},clip]{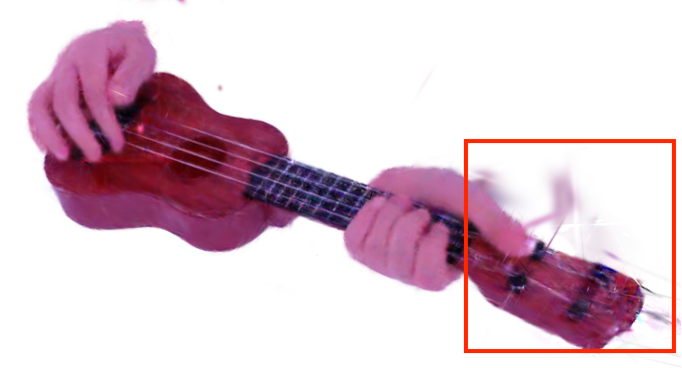}
            \\
	\end{tabular}
    \vspace{-0.1in}
	\caption{
		\label{Fig:NVS_comparison_1}
		\textbf{Qualitative Comparisons on Novel View Synthesis.} \methodname produces superior reconstruction quality with sharper novel view synthesis renderings. 
        Compared to baselines, \methodname delivers more fine-grained details, particularly in occluded regions and around edges, whereas baseline methods exhibit artifacts and blurriness.
        Please zoom in for better views.
	}
\vspace{-0.1in}

\end{figure*}

\begin{table*}[!t]
\small
\scalebox{1}{
    \begin{tabular}{p{2.35cm} |
>{\centering\arraybackslash}p{0.72cm} >{\centering\arraybackslash}p{0.72cm} 
>{\centering\arraybackslash}p{0.72cm} >{\centering\arraybackslash}p{1cm} |
>{\centering\arraybackslash}p{0.72cm} >{\centering\arraybackslash}p{0.72cm} 
>{\centering\arraybackslash}p{0.72cm} >{\centering\arraybackslash}p{1cm} |
>{\centering\arraybackslash}p{0.72cm} >{\centering\arraybackslash}p{0.72cm} 
>{\centering\arraybackslash}p{0.72cm} >{\centering\arraybackslash}p{1cm}
}

        Dataset & \multicolumn{4}{c|}{GigaHands}  & \multicolumn{4}{c|}{DiVa-360} & \multicolumn{4}{c}{MANUS-Grasps}\\
        Method\textbackslash{}Metric
        & SSIM$^\uparrow$   & PSNR$^\uparrow$    & LPIPS$^\downarrow$  & Mem$^\downarrow$ 
        & SSIM$^\uparrow$   & PSNR$^\uparrow$    & LPIPS$^\downarrow$  & Mem$^\downarrow$ 
        & SSIM$^\uparrow$   & PSNR$^\uparrow$    & LPIPS$^\downarrow$  & Mem$^\downarrow$ \\
        \hline 
        Deformable3DGS & \cellcolor{yellow!40}0.970 & 26.37 & \cellcolor{yellow!40}0.039 & 5MB & \cellcolor{yellow!40}0.978 & 29.20 & \cellcolor{yellow!40}0.036 & 5MB & \cellcolor{yellow!40}0.970 & \cellcolor{yellow!40}29.58 & \cellcolor{yellow!40}0.042 & 6MB \\
        4DGaussians & 0.963 & 25.82 & 0.045 & 136MB & 0.975 & 28.35 & 0.041 & 136MB & 0.967 & 28.60 & 0.044 & 136MB \\ 
        Realtime4DGS & 0.969 & 26.14 & 0.044 & 168MB & 0.957 & 21.99 & 0.062 & 103MB & \cellcolor{red!40}0.974 & \cellcolor{orange!40}29.85 & 0.044 & 291MB \\  
        AT-GS & \cellcolor{orange!40}0.972 & \cellcolor{yellow!40}26.62 & \cellcolor{orange!40}0.038 & 380MB & \cellcolor{orange!40}0.979 & \cellcolor{yellow!40}28.41 & \cellcolor{orange!40}0.033 & 160MB & \cellcolor{orange!40}0.972 & 29.40 & \cellcolor{orange!40}0.039 & 122MB \\
        3DGStream & 0.96 & \cellcolor{orange!40}28.12 & 0.061 & 15MB  & 0.960 & \cellcolor{orange!40}29.34 & 0.043 & 15MB & 0.946 & 26.41 &  0.084 & 16MB\\
        Ours& \cellcolor{red!40}0.982 &  \cellcolor{red!40}30.06 & \cellcolor{red!40}0.018 & 13MB & \cellcolor{red!40}0.983 & \cellcolor{red!40}32.18 & \cellcolor{red!40}0.020 & 11MB & \cellcolor{red!40}0.974 & \cellcolor{red!40}32.67 & \cellcolor{red!40}0.021 & 8MB\\
        
    \end{tabular}
    
}
\vspace{-0.1in}
\caption{
    \label{TAB:NVS_Comparison} {\textbf{Quantitative Comparisons on Dynamic Reconstruction.} We compare \methodname with other Gaussian-based approaches regarding novel view synthesis quality, demonstrating the superior performance of \methodname across SSIM, PSNR, and LPIPS metrics with efficient memory usage.}
}
\vspace{-0.1in}
\end{table*}

\subsection{Dynamic Reconstruction Datasets}
To evaluate dynamic reconstruction, we compare with baselines the performance of novel view synthesis on three public datasets: DiVa-360~\cite{lu2024diva}, MANUS-Grasps~\cite{pokhariya2024manus}, and GigaHands~\cite{fu2024gigahandsmassiveannotateddataset}, all of which provide high-quality, multi-view recordings of real-world tabletop manipulation tasks.
From \textit{DiVa-360}, we select $3$ daily-life, bi-manual tabletop manipulation sequences, each averaging $400$ frames.
We choose $20$ tabletop grasping sequences with an average of $200$ frames per sequence from \textit{MANUS-Grasps}~\cite{pokhariya2024manus}, and $8$ complex bimanual hand-object interaction sequences, each averaging $300$ frames from \textit{GigaHands}~\cite{fu2024gigahandsmassiveannotateddataset}.
In total, we evaluate dynamic reconstruction performance on $31$ sequences.
For each sequence, we choose around 30 camera views, with two views for testing and the remaining for training.
All metrics are calculated as the average over all testing frames.

\subsection{Baselines}
For dynamic scene reconstruction, we compare our method with five state-of-the-art approaches: 4DGaussians~\cite{wu20244d}, Deformable-3DGS~\cite{yang2024deformable}, Realtime4DGS~\cite{yang2023gs4d}, 3DGStream~\cite{sun20243dgstream}, and AT-GS~\cite{chen2024adaptive}. However, these methods do not naturally support dynamic contact estimation, as they do not differentiate between hands and objects as \methodname. Therefore, to evaluate contact estimation, we select another group of baselines that are state-of-the-art analytical methods for contact estimation: MANO~\cite{MANO:SIGGRAPHASIA:2017}, HARP~\cite{karunratanakul2023harp}, and MANUS~\cite{pokhariya2024manus}.

\subsection{Evaluation on Dynamic Contact Estimation}
\paragraph{Qualitative Comparisons.}
Figure \ref{Fig:contact_compare} presents qualitative comparisons of accumulated contact estimation on \datasetname benchmark against baseline methods. Our method produces more accurate and coherent dynamic contact maps that closely align with the ground truth, in contrast to the over-segmentation and noisy artifacts observed in other approaches. This improvement is attributed to three key factors:
(1) our template-based Gaussian representation enforces a strong inductive bias that reduces noise and removes floating artifacts around contact regions; (2) our deformation refinement module effectively captures time-dependent surface deformation; (3) our contact-guided adaptive density control strategy allocates additional Gaussian primitives to high-variation regions, allowing for fine-grained reconstruction of contact details.

\paragraph{Quantitative Comparisons.}
Table \ref{TAB:contact_comparison} quantitatively evaluates contact estimation accuracy using Intersection over Union (IoU) and F1-score metrics~\cite{pokhariya2024manus} by comparing estimated and ground truth contact maps. 
As shown in Table \ref{TAB:contact_comparison}, our method consistently outperforms all baselines on the \datasetname dataset, which aligns with the visual results in Figure \ref{Fig:contact_compare}.
Notably, while other Gaussian-based hand models\cite{pokhariya2024manus} use approximately 300k Gaussians per sequence, \methodname achieves superior results with only about 10k Gaussians per sequence on average, underscoring its efficiency in dynamic contact estimation.


\begin{figure}[t]
  \centering
  \includegraphics[width=0.98\linewidth]{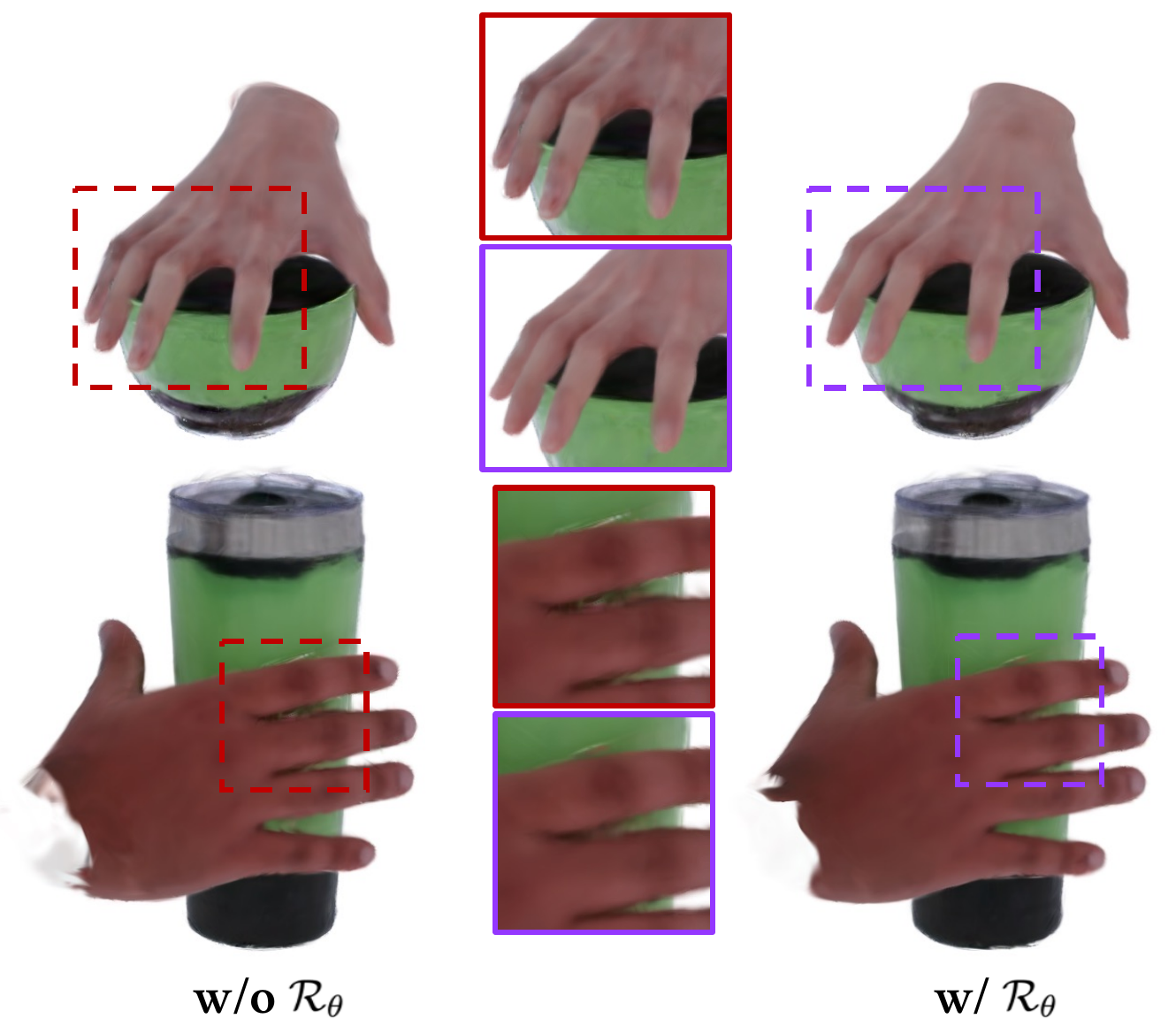}
  \vspace{-0.1in}
  \caption{\textbf{Ablation studies for the refinement module} $\mathcal{R}_{\theta}$. 
  $\mathcal{R}_{\theta}$ captures time-dependent deformation which reduces blurry artifacts around the contact region.
  }
  \vspace{-0.2in}
  \label{Fig:Ablation_MLP}
\end{figure}

\subsection{Evaluation on Dynamic Reconstruction}
\paragraph{Qualitative Comparisons.}
%
Figure~\ref{Fig:NVS_comparison_1} presents qualitative comparisons of novel view synthesis. 
Compared to baselines, our method presents higher reconstruction quality with clearer and more detailed reconstruction, especially in contact regions.
Our approach accurately reconstructs both low-frequency non-interactive areas and high-frequency hand–object interactions, thanks to our contact-guided adaptive density control strategy.
Moreover, while other unstructured Gaussian representations methods tend to generate floating artifacts, our template-based Gaussian representation enforces a strong inductive bias that reduces such artifacts with fewer Gaussian primitives, resulting in better 3D consistency.

%

\paragraph{Quantitative Comparisons.}
Table \ref{TAB:NVS_Comparison} presents quantitative comparisons on dynamic reconstruction with PSNR, SSIM and LPIPS~\cite{zhang2018unreasonable} metrics.
As indicated in Table~\ref{TAB:NVS_Comparison}, our method outperforms state-of-the-art baselines in all metrics across all scenes, demonstrating the superior performance and generalizability of \methodname. 
Notably, we achieve these improvements while maintaining fast optimization and efficient memory usage, underscoring the efficiency of our approach in modeling challenging hand–object manipulation scenes.

\subsection{Ablation Study}
\begin{figure}[t]
  \centering
  \includegraphics[width=\linewidth]{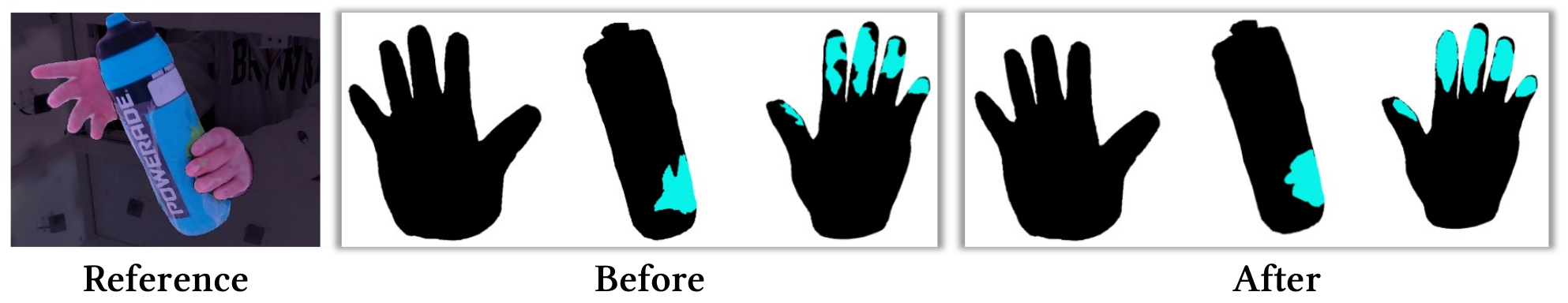}
  \vspace{-0.2in}
  \caption{\textbf{Ablation studies on the Contact-Guided Adaptive Density Control Strategy.} 
  This strategy effectively regulates the contact regions by allocating more isotropic gaussian surfels, yielding more accurate contact estimation.
  }
  \vspace{-0.1in}

\label{Fimanipulationsg:Ablation_1}
\end{figure}

%
\paragraph{Time-dependent Deformation Refinement Module.} Qualitative comparisons in Figure~\ref{Fig:Ablation_MLP} show that the refinement module $\mathcal{R}_{\theta}$ plays an important role in alleviating blurry artifacts around contacting regions caused by time-dependent high-frequency deformations of hand skin. 
We also observe consistent quantitative results in Table~\ref{tab:ablation_CGADCS} (\RomanNumeralCaps{2}) that removing $\mathcal{R}_{\theta}$ degrades the overall performance.
\paragraph{Contact-guided Adaptive Density Control Strategy.}
Figure~\ref{Fimanipulationsg:Ablation_1} shows that
the accuracy and coverage of contacts estimation improves by a large margin after introducing the contact-guided adaptive density control strategy.
Table.~\ref{tab:ablation_CGADCS} (\RomanNumeralCaps{1}) indicates that the contact-guided adaptive density control strategy helps accumulate more gradients among contacting regions and provide an effective inductive bias for the densification process during optimization, which is important to improve the accuracy of the occluded areas.

\begin{table}[t]
\small
\scalebox{1}{
    \begin{tabular}{p{1.35cm} |
>{\centering\arraybackslash}p{0.75cm}
>{\centering\arraybackslash}p{0.75cm} >{\centering\arraybackslash}p{0.75cm} 
>{\centering\arraybackslash}p{0.75cm} >{\centering\arraybackslash}p{1.25cm} 
}
         & mIoU$^\uparrow$  & SSIM$^\uparrow$   & PSNR$^\uparrow$    & LPIPS$^\downarrow$ & $\text{F}1$ score$^\uparrow$ \\
        \hline 
         \RomanNumeralCaps{1}. w/o CG & \cellcolor{yellow!40}0.216  & \cellcolor{orange!40}0.975 & \cellcolor{orange!40}31.82 & \cellcolor{orange!40}0.021 & \cellcolor{yellow!40}0.352 \\
         \RomanNumeralCaps{2}. w/o $\mathcal{R}_{\theta}$ & \cellcolor{orange!40}0.225  & \cellcolor{yellow!40}0.971 & \cellcolor{yellow!40}31.79 & \cellcolor{orange!40}0.021 & \cellcolor{orange!40}0.375 \\
         \RomanNumeralCaps{3}. Full & \cellcolor{red!40}0.226  & \cellcolor{red!40}0.978 & \cellcolor{red!40}31.88 & \cellcolor{red!40}0.020 & \cellcolor{red!40}0.378 \\ 
        
    \end{tabular}
}
  \vspace{-0.1in}
\caption{
    \label{tab:ablation_CGADCS} {\textbf{Ablation studies} for the refinement module $\mathcal{R}_{\theta}$ and the contact-guided adaptive density control strategy (CG). Both components improve reconstruction accuracy.}
}
\vspace{-0.2in}
\end{table}

%% file: sec/7_conclusion.tex
\section{Conclusion}
\label{Sec:Conclusion}
This paper introduces \methodname, a novel markerless method for capturing dynamic contacts in complex hand-object manipulations. 
The method utilizes a dynamic articulated representation based on 2D Gaussian surfels to effectively capture complex manipulations. 
\methodname takes advantage of the inductive biases from template models by binding surfels to MANO~\cite{MANO:SIGGRAPHASIA:2017} meshes, thereby efficiently stabilizing and accelerating the optimization process. 
A dedicated refinement module addresses time-dependent high-frequency deformations, while a contact-guided adaptive sampling strategy selectively refines surfel density around contacting regions.

To evaluate dynamic contacts, we have curated a new benchmark, \datasetname, which provides ground-truth accumulated contact data for $21$ complex and diverse real-world multi-view manipulation sequences. 
Extensive experiments demonstrate that \methodname achieves state-of-the-art accuracy in dynamic contact capture and significantly improves dynamic reconstruction quality, leading to high-fidelity novel view synthesis. 
